\documentclass{article}

\PassOptionsToPackage{numbers}{natbib}

\usepackage[main, final]{neurips_2026}

\usepackage[utf8]{inputenc} 
\usepackage[T1]{fontenc}    
\usepackage{hyperref}       
\usepackage{url}            
\usepackage{booktabs}       
\usepackage{amsfonts}       
\usepackage{nicefrac}       
\usepackage{microtype}      
\usepackage{xcolor}         

\usepackage{amsmath,amsfonts,bm}

\def\eqref#1{equation~\ref{#1}}

\def\1{\bm{1}}

\def\mA{{\bm{A}}}

\def\mE{{\bm{E}}}

\def\mH{{\bm{H}}}

\def\mQ{{\bm{Q}}}

\def\mS{{\bm{S}}}
\def\mT{{\bm{T}}}

\def\mW{{\bm{W}}}

\DeclareMathAlphabet{\mathsfit}{\encodingdefault}{\sfdefault}{m}{sl}
\SetMathAlphabet{\mathsfit}{bold}{\encodingdefault}{\sfdefault}{bx}{n}

\usepackage{wrapfig}
\usepackage{float}
\usepackage{algorithm}
\usepackage{algpseudocode} 
\usepackage[nameinlink]{cleveref}
\usepackage[english]{babel}
\definecolor{pycomment}{HTML}{5A8A4A}

\usepackage{graphicx}
\usepackage{tabularx}
\usepackage{array}
\usepackage{enumitem}

\newcommand{\method}{ASH}
\newcommand{\retmet}{ASH retrieval}
\newcommand{\pokemon}{Pok\'emon}
\newcommand{\Pokemon}{Pok\'emon}

\newcommand{\obs}{\ensuremath{\tau^{obs}}}
\newcommand{\lmem}{\ensuremath{\rho}}
\newcommand{\shortlen}{\ensuremath{w_s}}
\newcommand{\longlen}{\ensuremath{w_l}}
\newcommand{\retlen}{\ensuremath{w_r}}
\newcommand{\shorttok}{\ensuremath{\mT^s}}
\newcommand{\longtok}{\ensuremath{\mT^\ell}}
\newcommand{\DR}{\ensuremath{\mathcal{D}^R}}
\newcommand{\DI}{\ensuremath{\mathcal{D}^{I}}}
\newcommand{\Dpi}{\ensuremath{\mathcal{D}^{\pi}}}
\newcommand{\sample}{\ensuremath{\obs_{t:t+\shortlen}}}
\newcommand{\oneToN}{\ensuremath{1 \sim N}}

\title{ASH: Agents that Self-Hone in Long-Horizon Worlds}

\vspace{-1ex}
\author{%
  Benjamin Schneider\\
  University of Waterloo\\
  \texttt{benjamin.schneider@uwaterloo.ca} \\
  \And
  Xavier Schneider \\
  University of Waterloo \\
  \texttt{xschneider@uwaterloo.ca} \\
  \AND
  Victor Zhong \\
  University of Waterloo \\
  \texttt{victor.zhong@uwaterloo.ca} \\
  \And
  Sun Sun \\
  National Research Council Canada \\
  \texttt{sun.sun@nrc-cnrc.ca} 
}

\begin{document}

\maketitle

\vspace{-3ex}
\begin{abstract}
Long-horizon visuomotor tasks remain a fundamental challenge in AI, as current methods rely on hand-engineered rewards or action-labeled demonstrations, neither of which scales.
We introduce \textbf{\method{}}, an agentic system that learns a long-horizon policy from unlabeled, noisy internet video, without reward shaping or expert annotation. \method{} follows a self-improvement loop; when it gets stuck, \method{} learns an Inverse Dynamics Model (IDM) from its own trajectories, and uses its IDM to extract supervision from relevant internet video. \method{} uses unsupervised learning to identify key moments from large-scale internet video and retains them as long-term memory --- allowing it to tackle long-horizon problems.
We evaluate \method{} on two complementary environments demanding multi-hour planning: \emph{\Pokemon{} Emerald}, a turn-based RPG, and \emph{The Legend of Zelda: The Minish Cap}, a real-time action-adventure game.
In both games, behavioral cloning, retrieval-augmented and zero-shot foundation-model baselines plateau, while \method{} sustains progression across our 8-hour evaluation.
\method{} reaches an average of $11.2/12$ milestones in \Pokemon{} Emerald and $9.9/12$ in Legend of Zelda, while the strongest baseline gets stuck in both environments at an average of $9.3/12$ and $7.8/12$ milestones, respectively.
We demonstrate that self-improving agents are a scalable recipe for learning in long-horizon worlds.
Our project page is \href{https://toolbox-hq.github.io/ASH}{https://toolbox-hq.github.io/ASH}.
\end{abstract}

\section{Introduction}
\label{sec:intro}

Developing agents capable of mastering long-horizon, open-ended tasks remains a fundamental challenge in artificial intelligence~\citep{huang2026rethinkingmemorymechanismsfoundation,hu2024generalpurposerobotsfoundationmodels}.
Unlike agents for short-horizon planning, long-horizon agents must navigate complex, open-ended worlds over millions of steps. This requires multi-hour planning, fine-grained motor control, and the ability to adapt to new visual contexts.

Despite progress in visuomotor learning, current approaches struggle in these long-horizon settings.
Behavioral Cloning (BC) methods~\citep{ijcai2018p687} typically learn a policy from massive offline datasets.
At this scale, assembling high-quality
data is a significant challenge, requiring domain knowledge to
curate data and extensive human annotation~\citep{baker2022videopretrainingvptlearning}. Moreover, much of the available data may be irrelevant to the agent’s current objectives, increasing both training compute and susceptibility to noise.
Long-horizon tasks exacerbate these challenges~\citep{pmlr-v9-ross10a}: a single trajectory may span hours, making it intractable to model dependencies across the full sequence.
As a result, BC policies typically condition on short-term memory and struggle when a current decision depends on old information~\citep{lifshitz2023steve}.

While Reinforcement Learning (RL) has demonstrated success in complex environments, it remains heavily dependent on dense reward engineering~\citep{pleines2025pokemon,hafner2025trainingagentsinsidescalable,baker2022videopretrainingvptlearning}.
In a domain spanning tens of hours with diverse sub-tasks, manually designing a reward function for every intermediate
milestone is brittle, domain-specific, and labor intensive. Like BC, modeling dependencies across hour-long sequences using RL is intractable. Instead,
a common approach is to use short-term memory combined
with ``next sub-objective'' conditioning~\citep{hafner2025trainingagentsinsidescalable,baker2022videopretrainingvptlearning}, which amounts to a human manually laying out a path for the agent to follow. These approaches have difficulty learning skills and knowledge beyond their human-engineered objectives, making them unsuitable for open-ended environments.

\begin{figure*}[t!]
\centering
\includegraphics[width=\linewidth,trim={0 0 0 0},clip]{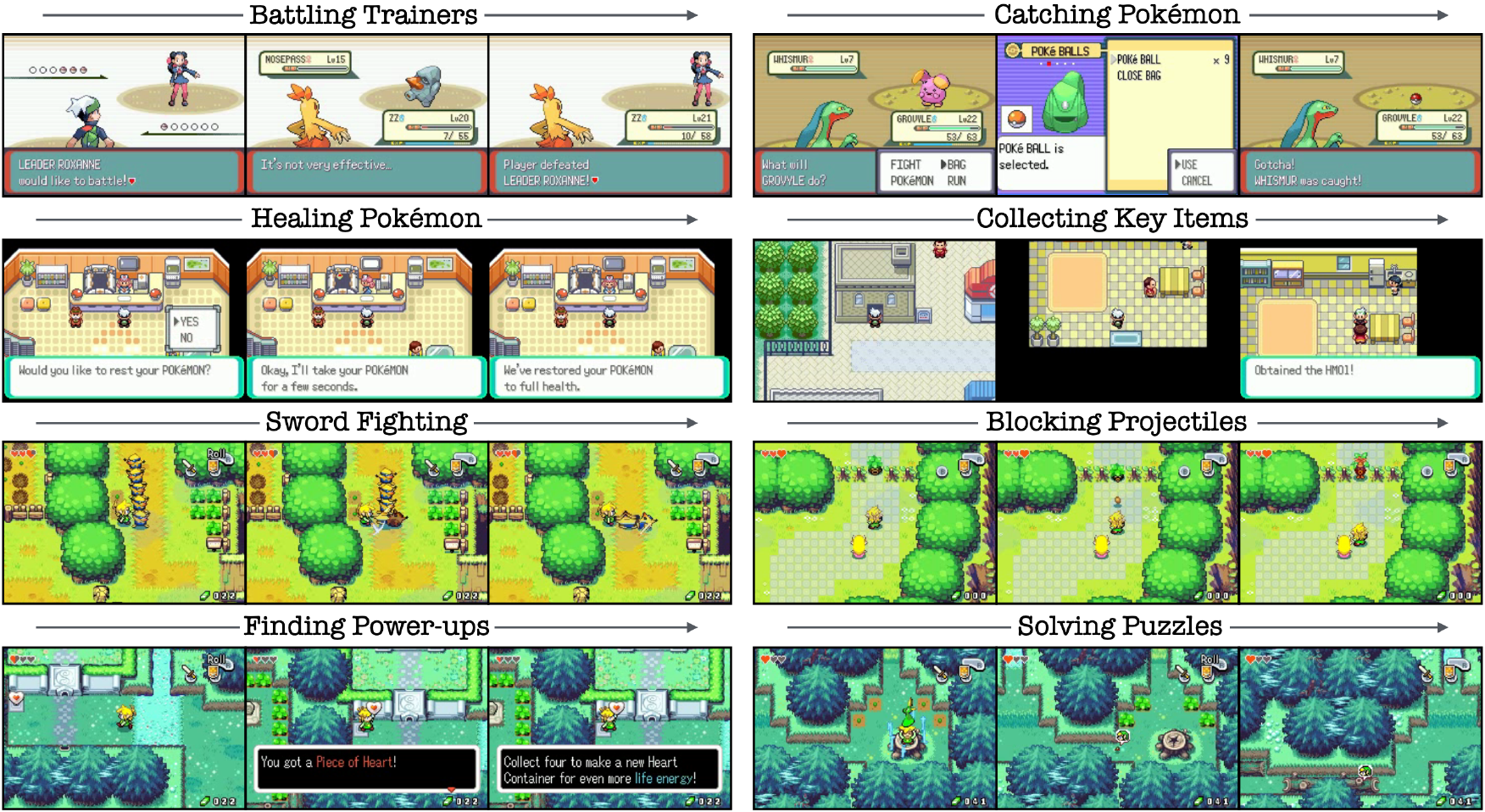}
\vspace{-3ex}
\caption{\method{} self-improves over the course of a multi-hour playthrough by retrieving and learning from relevant internet video. In \pokemon{} Emerald, \method{} learns to battle trainers, catch \pokemon{}, and heal its team at \pokemon{} centers. In Zelda, \method{} discovers how to wield a sword and apply game mechanics to solve puzzles. These skills compound, allowing \method{} to achieve sustained progression.}
\label{fig:imag}
\vspace{-3ex}
\end{figure*}

To address these limitations, we introduce \textbf{\method{}}, an agentic system that follows a \textbf{dynamic bootstrap} loop. When \method{} gets stuck, it searches for helpful demonstrations on the internet, and learns from them. \method{} uses its current trajectories to retrieve relevant examples from online video and to update its Inverse Dynamics Model (IDM). \method{} uses unsupervised learning to identify key moments from the retrieved video. \method{} bootstraps its policy using these key moments as \textbf{long-term memory} and the retrieved video demonstrations labeled by the IDM.

We evaluate \method{} in two complementary long-horizon environments: \emph{\Pokemon{} Emerald}, a turn-based RPG combining strategic battling with open-ended exploration, and \emph{The Legend of Zelda: The Minish Cap}, a real-time action-adventure game that requires puzzle solving and fine-grained motor control.
Both require dozens of hours of decision-making across millions of distinct visual states, which makes them useful as long-horizon testbeds.
\method{} interacts with the games as a human does: by observing the screen and pressing buttons.

In our evaluations, \method{} progresses further than behavioral cloning~\citep{baker2022videopretrainingvptlearning}, retrieval-augmented~\citep{10655182}, and zero-shot foundation-model~\citep{qwen35blog} baselines.
Baselines plateau on intermediate milestones, while \method{} continues to acquire new skills---battling trainers, catching \pokemon{}, exploring areas, and defeating enemies---across multi-hour horizons.
After 8 hours, \method{} achieves an average of $11.2/12$ progression milestones in \Pokemon{} Emerald and $9.9/12$ milestones in Legend of Zelda.
The strongest baseline gets stuck in both environments, achieving an average of $9.3$ and $7.8$ milestones, respectively.

\method{}'s main contributions are as follows: \textbf{(1)} We introduce {dynamic bootstrapping for long-horizon learning}. \textbf{(2)} We propose {long-term memory} created from salient moments in large-scale internet video. \textbf{(3)} Through experimentation, we demonstrate that dynamic bootstrapping and long-term memory create a scalable recipe for long-horizon learning --- capable of multi-hour planning, fine-grained motor control, and the ability to adapt to new visual contexts.

\section{Related Work}
\label{sec:related}

\textbf{Internet-scale behavioral cloning.}
A large body of work trains visuomotor policies via BC on observation-only video at scale.
VPT~\citep{baker2022videopretrainingvptlearning} creates pseudo-actions for thousands of hours of Minecraft video with an IDM trained on contractor-collected gameplay~\citep{ijcai2018p687}, and remains the strongest reported Minecraft BC backbone~\citep{li2025optimus2}.
A parallel \emph{imitation-from-observation} literature recovers actions without an explicit IDM~\citep{torabi2018behavioralcloningobservation, pmlr-v97-edwards19a, lapo, pmlr-v235-bruce24a}.
\method{} departs from these approaches in two ways: it does not rely on a single statically curated training corpus or a fixed action-labeling model. Instead, \method{} is an agentic system that uses the trajectories of its agents to update its IDM and retrieve data for policy bootstrapping.

\textbf{Retrieval-augmented decision-making.}
Retrieval has been used to give agents relevant context~\citep{lewis2020rag,yao2022react,shinn2023reflexion}, to condition policies on prior experience~\citep{humphreys2022largescale,pmlr-v162-goyal22a}, and to guide planning~\citep{zhang2023large}.
In these methods, retrieval selects a small number of relevant documents, trajectories, or examples to condition the agent during inference. \method{} uses retrieval differently: it retrieves a task-relevant training set from a much larger internet-scale corpus, then updates its weights using that data.

\textbf{Self-improvement from model-generated data.}
A growing line of work bootstraps a learner from data produced by an earlier version of itself: STaR~\citep{zelikman2022star} and ReST~\citep{gulcehre2023reinforcedselftrainingrestlanguage} for reasoning, self-play for games~\citep{10.5555/3295222.3295288,silver2017masteringchessshogiselfplay}, and self-distillation in robotics~\citep{gdm2024autort}.
\method{} fits this template, but its improvement comes from retrieved internet video rather than from an external verifier or reward model, which are often unavailable in open-ended environments~\citep{fan2022minedojo}.

\textbf{Long-horizon control in visuomotor agents.}
Long-horizon control requires agents to retain information that may become useful much later. However, full-history conditioning is computationally intractable, while short fixed-length memory is efficient but can lose information needed for planning~\citep{lifshitz2023steve}. BC and RL are dominated by bounded-memory policies, including recurrent policies that compress history into a fixed-size state and transformers that condition on a fixed context window~\citep{chen2021decisiontransformerreinforcementlearning,baker2022videopretrainingvptlearning,hafner2025trainingagentsinsidescalable}.
Memory-augmented language architectures~\citep{wu2022memorizing,bulatov2022recurrent,pmlr-v162-borgeaud22a} and skill-library agents such as Voyager~\citep{wang2023voyageropenendedembodiedagent} offer alternatives.
Rather than being authored by an LLM or hand-engineered, \method{} autonomously discovers key moments from internet data, and retains those moments as long-term memory when it encounters them in the environment.

\textbf{RL in open-ended environments.}
RL can achieve strong results when dense rewards, self-play objectives, or programmatic verifiers are available~\citep{mnih2013playingatarideepreinforcement,silver2017masteringchessshogiselfplay,Vinyals2019GrandmasterLI,openai2019dota2largescale,hafner2025trainingagentsinsidescalable}. However, these assumptions are hard to satisfy in open-ended environments. For example, \citet{pleines2025pokemon} train a PPO agent on \Pokemon{} Red by manually creating a trail of rewards for the agent to follow. They report high failure rates when the agent encounters situations not anticipated by the reward engineer, and must continually adapt their rewards to correct for these failure modes. By contrast, \method{}'s training signal comes from unlabeled demonstrations, a scalable approach to open-ended long-horizon learning.

\textbf{Foundation-model agents and scaffolded play.}
A separate line of work uses pretrained foundation models with extensive scaffolding as agents.
Voyager~\citep{wang2023voyageropenendedembodiedagent} uses an LLM and a learned skill library in a Minecraft code-as-action loop, and \citet{karten2025pokeagent,twitchClaudePlaysPokemonTwitch} apply pretrained models to \Pokemon{} Emerald with scaffolds that read underlying game memory.
These systems demonstrate emergent behavior but rely on privileged state access to make the games interpretable by an LLM~\citep{simateam2025sima2generalistembodied}.
Unlike these approaches, \method{} offers a path to long-horizon learning beyond environments where privileged state is available.

\textbf{Long-horizon games as testbeds.}
Prior testbeds for long-horizon learning include Minecraft~\citep{baker2022videopretrainingvptlearning,hafner2025trainingagentsinsidescalable,wang2023voyageropenendedembodiedagent,fan2022minedojo}, NetHack~\citep{hambro2023dungeonsdatalargescalenethack}, and \Pokemon{}~\citep{pleines2025pokemon,karten2025pokeagent}.
We add \emph{The Legend of Zelda: The Minish Cap} as a complement to \Pokemon{} Emerald. Zelda contains real-time action and puzzle solving, while still offering long-horizon objectives. This contrasts with the turn-based, strategic nature of \Pokemon{}.

\section{\method}
\label{sec:method}

\textbf{Preliminary.}
An agent interacts with an environment $\mathcal{E}$ that, given action $a \in \mathcal{A}$, returns observation $o \in \mathcal{O}$. We write \emph{observation-action trajectories} as $\tau = (o_1, a_1, \dots, o_T, a_T)$ and \emph{observation-only trajectories} as $\obs = (o_1, \dots, o_T)$.
A policy specifies a distribution over the agent's next action given the information available to it, written $\pi \sim p_\theta(a_{t} \mid \cdot)$. We use end-exclusive slicing $A_{i:j} = (A_i, \dots, A_{j-1})$. A detailed symbol glossary appears in \Cref{apx:notation}.

Behavioral cloning (BC)~\citep{ijcai2018p687} trains $\pi$ via supervised learning using cross-entropy loss on observation-action trajectories. However, scalable observation-only sources such as YouTube~\citep{baker2022videopretrainingvptlearning} lack action labels. We use an IDM to supply pseudo-actions from these observation-only trajectories; $\text{IDM}(\obs)=(\tilde{a}_1, \tilde{a}_2, \dots, \tilde{a}_T)$.
The IDM is trained via supervised learning on a dataset of observation-action trajectories.
We note that training an IDM is strictly easier than training a policy, as the IDM can use future information to predict pseudo-actions~\citep{baker2022videopretrainingvptlearning}.

We use HDBSCAN~\citep{hbscan} clustering to discover recurring key moments because (1) the number of clusters (key moments) is unknown a priori, (2) it rejects outliers as noise rather than forcing assignment, and (3) HDBSCAN's approximate-predict supports online assignment without refitting.

\begin{wrapfigure}[24]{R}{0.45\textwidth}
\vspace*{-5ex}\hfill%
\begin{minipage}{\linewidth}
\input{algorithms/ash_intro}
\end{minipage}
\end{wrapfigure}
\textbf{Overview.}
\method{} is analogous to a human that plays the game, gets stuck, and finds helpful video on the internet. We deploy $N$ agents under a single shared policy $\pi$ conditioned on short- and long-term memory, and collect trajectories $\tau_{\oneToN}$. A key-moment classifier $\mathcal{K}(\obs, t)$, derived from internet data, is used to determine if the observation at $t$ is a new key moment. When an observation is classified as a new key moment, it gets added to an agent's long-term memory and resets a per-agent stuck-timer. When any timer exceeds threshold $\Delta$, \method{} \emph{dynamically bootstraps}: it uses the agents' trajectories to update the IDM, and to retrieve visually similar internet video. These observation-only videos are used to update $\mathcal{K}$. The IDM then labels the videos with pseudo-actions, producing trajectories that are used to update $\pi$. Full \method{} pseudo-code can be found in \Cref{apx:ash}.


\textbf{Architecture.}
In the long-horizon setting, a decision can hinge on information from many steps ago. \method{} resolves this with a \emph{dual memory}: a short-term memory of the $\shortlen$ most recent observation-action pairs for reactive control, and a long-term memory of observations $\lmem$ that are the $\longlen$ most recent \emph{key moments}. The policy is therefore $\pi \sim p_\theta\!\left(a_{t} \mid \tau_{t-\shortlen:t},\, \lmem\right)$. \method{} uses an image tokenizer $\phi$ (SigLIP~\citep{zhai2023sigmoidlosslanguageimage}), an IDM with VPT's bidirectional-attention architecture~\citep{baker2022videopretrainingvptlearning}, and the causal transformer block architecture from \citet{yang2025qwen3technicalreport}, repeated 28 times, for $\mathcal{T}$.

\begin{wrapfigure}[13]{R}{0.45\textwidth}
\vspace*{-5ex}\hfill%
\begin{minipage}{\linewidth}
\input{algorithms/inference}
\end{minipage}
\end{wrapfigure}

\textbf{Training.}
We train the policy $\pi$ via gradient descent on observation-only trajectories $\obs$ of internet data. Each $\obs$ is divided into non-overlapping samples of length $\shortlen$, denoted $\sample = (o_{t}, o_{t+1}, \dots, o_{t + \shortlen - 1})$.

\noindent
$\displaystyle
\begin{gathered}
\tilde{\mA} = \text{IDM}(\sample), \qquad \shorttok, \longtok = \phi(\sample, \lmem) \\[1ex]
\mH = \mathcal{T}\!\left(\left[\longtok,\; \shorttok_1, \tilde{\mA}_1,\; \dots,\; \shorttok_{\shortlen}, \tilde{\mA}_{\shortlen}\right]\right) \\[1ex]
\mathcal{L} = -\sum_{j=0}^{\shortlen} \tilde{\mA}_j^\top \log \mathrm{softmax}\!\left(\mW_o \mH_{\longlen + 2j + 1}\right).
\end{gathered}$

First, we use the IDM to create pseudo-actions for each observation in the sample $\sample$, which are encoded as one-hot vectors. These actions are used for both teacher-forcing and output labels while training $\pi$.
$\phi$ is used to tokenize short-term ($\sample$) and long-term ($\lmem$) observations to create $\shorttok$ and $\longtok$, respectively.
The resulting tokens are concatenated into a unified sequence where the long-term memories $\longtok$ act as a prefix to the interleaved short-term observations and pseudo-actions. This combined sequence is passed through the causal transformer $\mathcal{T}$, allowing the model to attend to important memories while predicting local actions.
We use linear layers to project $\shorttok, \longtok$ and $\tilde\mA$ into the hidden dimension of $\mathcal{T}$.
The output layer $\mW_o$ maps the hidden output at the observation token position of the transformer $\mH$ to the action classes.
The policy $\pi$ is trained using $\mathcal{L}$, the cross-entropy loss between predictions generated by \method{} and pseudo-actions generated by our IDM.


\textbf{Inference.}
\Cref{alg:agent_inference} describes \method's inference loop. Each agent $n$ maintains a per-agent stuck timer $C_{n}$, initialized to $0$, that counts how many consecutive steps have elapsed since the agent last observed a new key moment. At each step, agent $n$ reads the $\longlen$ most recent key moments from its memory bank $\mathcal{M}_n$ as $\lmem$ (long-memory), combines them with the last $\shortlen$ observation-action pairs in $\tau_n$ (short-memory), and samples an action from $\pi$. After executing the action, the agent appends the resulting observation-action pair to $\tau_n$. The classifier $\mathcal{K}$ determines whether the observation is a new key moment. If so, the observation is appended to $\mathcal{M}_n$ and the timer $C_{n}$ is reset to zero; otherwise $C_{n}$ is incremented. The loop runs until any agent's timer exceeds $\Delta$, at which point \method{} treats that agent as stuck, halts inference for all agents, and bootstraps their shared policy. After the bootstrap, agents resume inference from their current observation.


\begin{wrapfigure}[29]{R}{0.45\textwidth}
\vspace*{-5ex}\hfill%
\begin{minipage}{\linewidth}
\input{algorithms/retv2.tex}
\end{minipage}
\vspace{-3ex}
\end{wrapfigure}

\textbf{Retrieval.}
When an agent gets stuck, \method{} needs training data targeted at its \emph{current roadblock}. To find this data, \method{} retrieves a dataset $\DR$ from internet video $\DI$ by selecting videos that are similar to the agent's current observations.
We index our internet dataset using DINOv2~\citep{oquab2024dinov2learningrobustvisual}
by embedding video frames sampled at $2$ second intervals.
Each internet video is represented as a sequence of L2-normalized visual embeddings, forming a matrix $\mE \in \mathbb{R}^{e \times d}$, where $e$ denotes the number of sampled frames and $d$ is the embedding dimension.
For each agent trajectory $\tau_n$, we similarly compute a trajectory embedding matrix $\mQ \in \mathbb{R}^{q \times d}$.
For each internet video $\tau^{\mathrm{obs}}$, we compute a similarity matrix
$\mS = \mQ \mE^\top$.
Retrieval searches for the most similar contiguous segment of length $\retlen$ in each internet video.
We score each temporal window $\mS_{:,\,t:t+\retlen}$ by greedy one-to-one matching --- each internet video frame in the window claims its highest-similarity trajectory frame, and each trajectory frame is consumed at most once. An internet video's score is its best temporal window score. We retrieve the top-$k$ videos per agent trajectory and union them into $\DR$. This matching scheme is compared to simpler baselines in \Cref{exp:ret}.

\textbf{Discovering key moments.}
$\mathcal{K}$ is built by applying the HDBSCAN~\citep{hbscan} clustering algorithm over the embeddings of $\DR$, and all previously clustered embeddings.
We further filter the resulting clusters by requiring each cluster contain observations from a minimum of $c_{\min}$ distinct trajectories in $\DR$.
This ensures that the clusters capture key visual moments that are commonly occurring between many trajectories.
Each individual cluster therefore aims to be a \emph{single key moment} in the environment.
Details on cluster quality can be found in \Cref{exp:clustering}.

\textbf{Key moments to long-term memory.}
$\mathcal{K}(\tau^{obs}, t)$ is a binary classifier: it returns \emph{true} if the observation at timestep $t$ is a \emph{new} key moment---one whose corresponding cluster has not yet been matched by any earlier observation in $\tau^{obs}$---and \emph{false} otherwise. This is done by embedding $\tau^{obs}_{t}$ using DINOv2~\citep{oquab2024dinov2learningrobustvisual} and applying HDBSCAN's approximate predict function to determine if it belongs to a cluster. These new key moments serve a dual purpose:
firstly, they are appended to an agent's memory bank $\mathcal{M}_n$, and the last $\longlen$ entries are used as long-term memory for the agent.
Secondly, they are used for our dynamic bootstrapping algorithm, where a bootstrap is triggered when an agent does not encounter a new key moment for $\Delta$ steps.


\textbf{Bootstrapping.}
Each bootstrap~(\Cref{alg:agent_bootstrap}) updates $\mathcal{K}$, the IDM and $\pi$. First, $\mathcal{K}$ is updated using the newly retrieved $\DR$. For IDM data, \method{} uses the observations from $\tau_{\oneToN}$ as IDM inputs, and the corresponding actions as labels. To ensure coverage of new environmental dynamics, we supplement the IDM training dataset with samples obtained from the environment using a random policy.
\pagebreak
\begin{wrapfigure}[16]{r}{0.45\textwidth}
\vspace*{-3ex}\hfill%
\begin{minipage}{\linewidth}
\input{algorithms/bootstrap}
\end{minipage}
\vspace{-3ex}
\end{wrapfigure}
\method{} uses $\mathcal{K}$ and the IDM to update the policy $\pi$. Since $\obs$ is internet video, it only contains a sequence of observations, not actions. Therefore, we use the IDM to create pseudo-actions for each sample from $\obs$. Then, $\mathcal{K}(\tau^{obs}, k)$ is used to construct memories ($\lmem$) from timesteps $k < t$---before the start of the sample---which are used for training $\pi$'s long memory. The resulting sample, consisting of $\shortlen$ observation-action pairs and $\longlen$ long-term memories, is added to $\pi$'s training dataset $\Dpi$. $\pi$ is then trained using $\Dpi$. For both $\pi$ and the IDM, we use 10\% of samples as a held-out set to prevent overfitting. Once $\pi$ has been bootstrapped, \method{} resumes agent inference.
Through this self-improvement loop, \method{} 
hones its policy---adapting to new aspects of the environment and refining the skills it has acquired.

\begin{figure*}[!t]
\centering
\vspace*{-2ex}
\includegraphics[width=\linewidth]{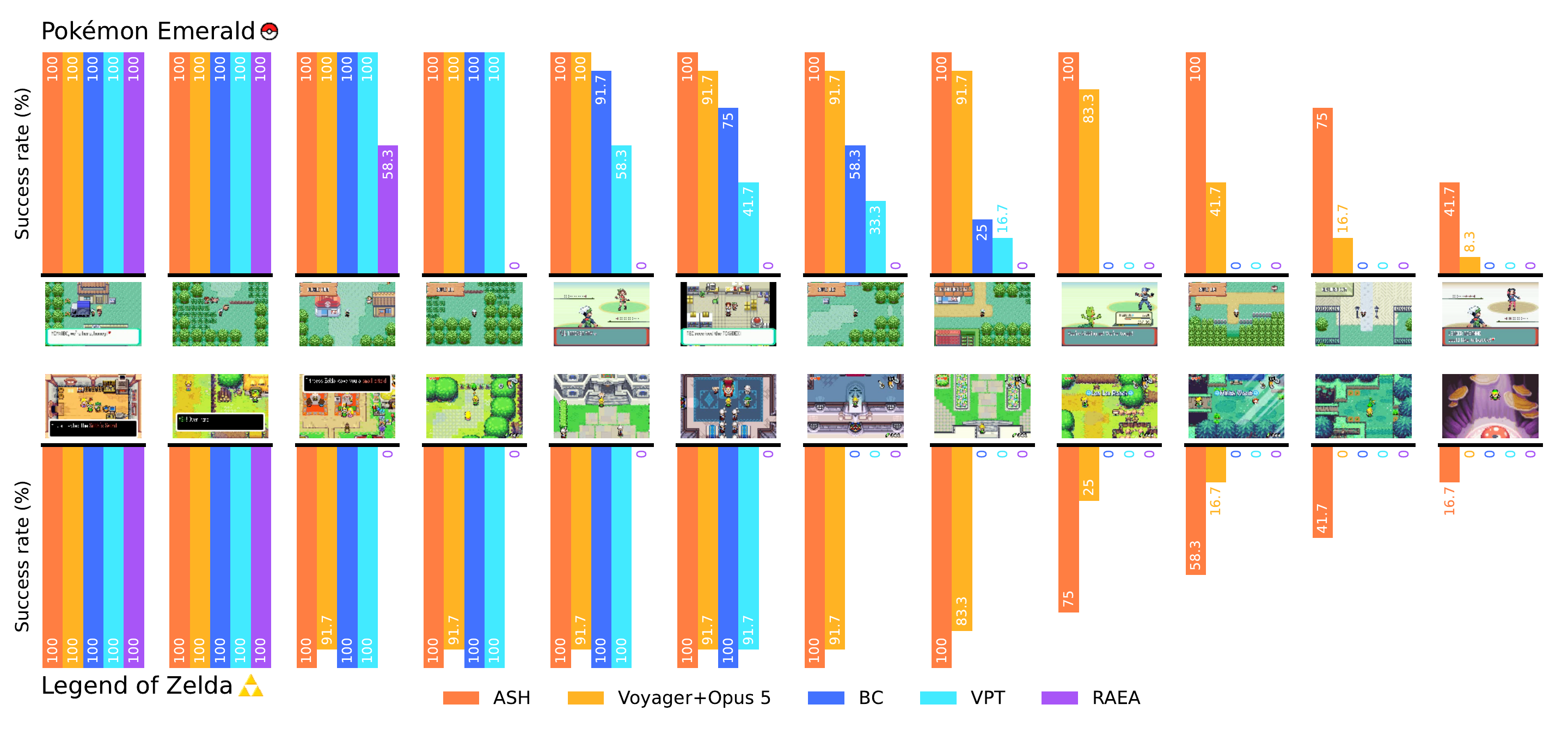}
\vspace*{-3ex}
\caption{Agent progress in \textbf{\Pokemon{} Emerald} (top) and \textbf{Legend of Zelda} (bottom), measured using milestone completion rates.
\method{} is able to adapt and continue progressing throughout the 8-hour gameplay period.
While all methods are able to complete early milestones, only \method{} can adapt to new areas, objectives, and mechanics. See \Cref{apx:final_milestones} for standard deviation.
}
\vspace*{-3ex}
\label{fig:ash}
\end{figure*}

\section{Experiments}
\label{experiments}

We evaluate \method{} on two complementary long-horizon environments:
\Pokemon{} Emerald, a strategic turn-based RPG with a deep skill tree and dozens of hours of content, and 
\emph{The Legend of Zelda: The Minish Cap}, an action-adventure game that demands real-time dexterity and puzzle solving.
Both span millions of distinct environment observations. We provide additional background on our environments in \Cref{apx:pokemon}.
Through our experiments, we investigate the following:

\begin{itemize}[
    leftmargin=1.2em,
    itemsep=1pt,
    topsep=1pt,
    parsep=0pt,
    partopsep=0pt
]
    \item How does \method{} compare to other adaptive and non-adaptive baselines?~(\Cref{exp:main})
    \item What effect do long-term memory and dynamic bootstrapping have on \method{}'s policy?~(\Cref{sec:ablation})
    \item What does bootstrapping do to the IDM?~(\Cref{exp:idm})
    \item Does \method{} retrieve relevant, high-quality examples from internet data?~(\Cref{exp:ret})
    \item Does \method{}'s long-term memory improve long-horizon decision-making?~(\Cref{exp:clustering})
    \item Can \method{}'s final policy replay the environment offline?~(\Cref{exp:forget})
\end{itemize}

\textbf{Milestones.} We evaluate our agents using \emph{milestones}, which are human-defined markers of progress in our environments. We derive our milestones from online walkthroughs~\citep{Bulbapedia_2025,minishcap}, descriptions of each milestone can be found in \Cref{apx:milestones}.

\textbf{Settings and implementation details.} In our experiments, \method{} uses $N=4$ agents.
We use a timestep of $0.25$ seconds to sample observations from the environment and input actions.
We set \shortlen{} and \longlen{} to $64$ and $10$, respectively.
At each timestep, we use \shortlen{} previous observations, with a sampling interval of $0.5$ seconds, as an agent's short-term memory. This allows the agent to maintain a short-term memory of $32$ seconds, while interacting with the environment every $0.25$ seconds.  
For retrieval, we set $k=100$ and $\retlen{}=540$.
We set our HDBSCAN~\citep{raschka2020machine} hyperparameter, $c_{\min}$, to $30$.
We scrape our internet datasets (\DI) from YouTube using a list of keywords~(\Cref{apx:keyword}).
Our dataset for \Pokemon{} Emerald contains approximately $22,000$ unique videos, whereas our Legend of Zelda dataset contains $17,000$ videos.
For dynamic bootstrapping, we set $\Delta$ to $4800$ which is $20$ minutes of real-time (the median video length in \DI).
\method{} contains $1.2$ billion parameters, $100$ million for vision encoder $\phi$ and $1.1$ billion for the transformer $\mathcal{T}$.
We use fp32 precision for $\phi$ and bf16 for $\mathcal{T}$.
We find that initializing each run from random weights is inefficient, so we use checkpoints trained for 3 and 9 GPU hours to initialize the IDM and $\pi$.
The data for these short training runs is obtained from the environment's starting state using a uniformly random policy over the action space (for the IDM) and \Cref{alg:agent_retrieval} (for $\pi$).
We detail architectural, inference, and training hyperparameters for \method{} in \Cref{apx:settings}.

\textbf{Baselines.}
We compare \method{} with the following baselines.
\textbf{Voyager+Opus 5}~\citep{wang2023voyageropenendedembodiedagent}: We scaffold Opus 5~\citep{anthropic2026claudeopus5} with a self-editable code skill library. Voyager+Opus 5 interacts with the game by sending button press actions and receiving frames as observations.
\textbf{VPT}~\citep{baker2022videopretrainingvptlearning}: VPT requires an existing IDM, so we use the strongest IDM checkpoint produced by \method{}, as measured by performance on a validation set; see \Cref{exp:idm}. We find that direct offline training on \DI{} creates a weak policy that does not progress past the first milestones, so we manually curate 400 hours of high-quality training data.
\textbf{Offline BC}: \method{}'s architecture without long-term memory or bootstrapping, trained using the same IDM and data as VPT.
\textbf{Retrieval-augmented Embodied Agent (RAEA)}~\citep{10655182}: We follow the recipe of ~\citet{10655182} and augment offline BC with cross-attention on retrieved examples. We split our manually curated data into 350 hours of training data and 50 hours of retrievable examples, labeled by our IDM. For both inference and training, we run \method{} retrieval every $32$ seconds with $\retlen{}=16$ and $k=1$ but return the best matching segment, rather than the whole video. Therefore, at every step, RAEA is conditioned on $32$ seconds of context and a $32$ second policy example. For compute parity, we allocate up to $288$ GPU hours ($4$ days on $4{\times}$H100 SXM 80GB GPUs) total for each method, or until overfitting is detected on held-out data.

\vspace{-1ex}
\subsection{Performance on Milestones}
\label{exp:main}\label{sec:lz}

For our baselines, we train one policy for each environment and perform 12 runs each. Since \method{} infers multiple agents at once, we perform 3 training runs per environment with $N=4$. This gives us 12 trajectories per-method. We report average milestone completion rate in \Cref{fig:ash}. We find that:

\begin{wrapfigure}[18]{r}{0.25\columnwidth}
\vspace{-3ex}
\centering
\includegraphics[width=\linewidth]{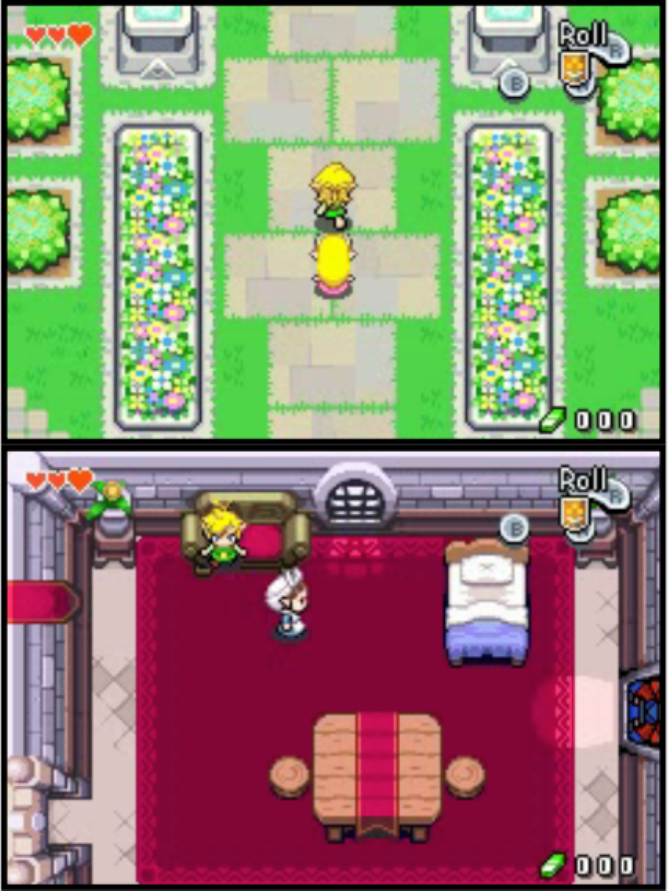}
\vspace{-3ex}
\caption{Outside (top) vs.\ inside (bottom) of the Zelda castle: offline policies collapse on this dynamics shift; \method{} bootstraps and continues.}
\label{fig:dynamics}
\vspace{-2ex}
\end{wrapfigure}

\textbf{Self-improvement is necessary for sustained progression.}
Over the 8-hour evaluation, \method{} reaches milestone 12 in both environments, while no baseline exceeds milestone 8 in \pokemon{} or 6 in Zelda. VPT and offline BC plateau once the games introduce dynamics that are underrepresented in their training distribution (for example, seldom encountered random events, or areas that are only briefly visited by players). RAEA, designed for short-horizon tasks, struggles in our long-horizon setting since it retrieves a different policy example every $32$ seconds, causing instability. Finally, Voyager+Opus 5 often loses track of its current objective, causing it to get stuck attempting the wrong objectives.

\textbf{\method{} adapts to changing dynamics.}
The Zelda castle entrance is the cleanest example of a distribution-shift moment (\Cref{fig:dynamics}): the visual dynamics change abruptly, and VPT and offline BC, frozen at training time, devolve into random button-mashing and never exit the castle. \method's policy also falters, but after a bootstrap is able to continue making progress. The same pattern appears in \Pokemon{} the moment the agent leaves the starting town and encounters the \pokemon{} battling interface.

\textbf{Bootstrapping evolves new skills.}
\method{}'s initial checkpoint does not exhibit any ability to battle with \pokemon{}, and therefore gets stuck on the second milestone (which requires battling).
However, through bootstrapping, \method{} becomes an adept battler. After hour $4$, \method{} wins $87\%$ of the \pokemon{} battles it engages in. 
In the Legend of Zelda, \method{} kills enemies with its sword in $75\%$ of trajectories. This capability is absent from its initial checkpoint and is not observed in any of our baselines.
We provide examples of \method{} evolving skills in \Cref{apx:playthrough}.

Across both environments, gameplay has a horizon beyond which a fixed policy struggles to succeed: the environments continually introduce new mechanics, regions, and visual contexts. Methods without an update mechanism plateau at that horizon. \method{}'s self-improvement loop allows it to move beyond this horizon.

\begin{figure*}[!t]
\vspace{-2.5ex}
\centering
\begin{minipage}{0.49\textwidth}
\centering
\includegraphics[width=\linewidth]{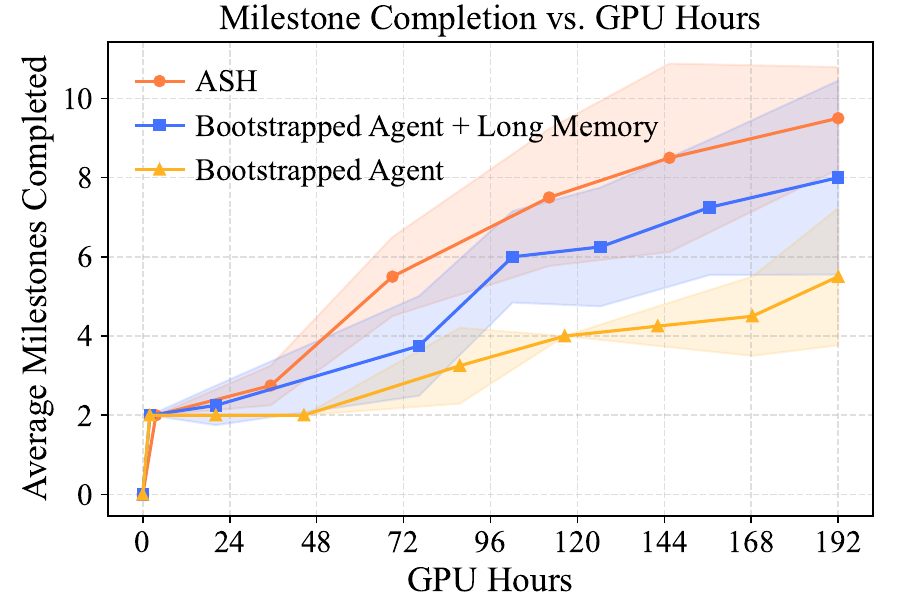}
\end{minipage}
\hfill
\begin{minipage}{0.49\textwidth}
\centering
\includegraphics[width=\linewidth]{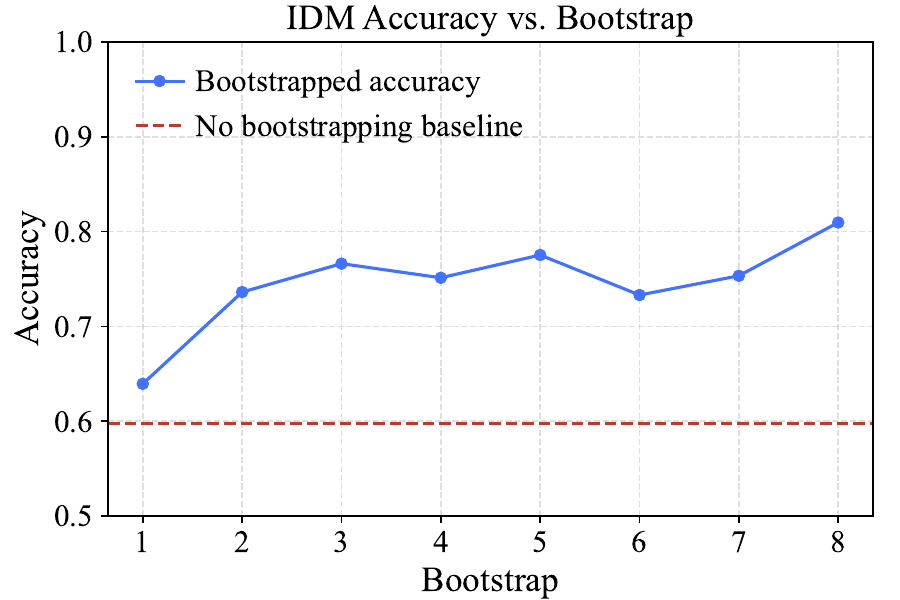}
\end{minipage}
\vspace{-0.5ex}
\caption{\textbf{(Left)} Component ablation on \Pokemon{} Emerald: each addition (long-term memory, dynamic bootstrapping) yields a clear gain in milestones completed per GPU hour of online training. Shaded regions are one standard deviation over $4$ trajectories per method. \textbf{(Right)} IDM accuracy across bootstraps, evaluated on a test set. The dashed line is the pre-bootstrap initialization checkpoint.}
\label{fig:ablation}\label{fig:idm}
\vspace{-2ex}
\end{figure*}

\subsection{Component Ablations}
\label{sec:ablation}
In \Cref{fig:ablation} (left), we ablate across the components of \method{} and measure their contribution to milestone completion in \Pokemon{} Emerald. \textbf{Bootstrapped Agent} removes both the long-term memory and dynamic bootstrapping, fixing the bootstrap interval to $\Delta$. \textbf{Bootstrapped Agent + Long Memory} reintroduces the agent's long-term memory but keeps the fixed bootstrapping schedule. \textbf{\method{}} adds dynamic bootstrapping. Because each variant uses different amounts of compute for inference and training, we plot milestone completion against \emph{total GPU hours of online training}. Each method is averaged over 4 trajectories, and the shaded regions show the standard
deviation. We find that the introduction of each component
of ASH results in a clear increase in performance. By the end of the evaluation, ASH’s long-memory component contributes 2.5 average milestones. Finally, dynamic bootstrapping allows \method{} to deploy compute more efficiently by only training when it gets stuck, contributing an additional 1.5 milestones.

\subsection{IDM Adaptation}
\label{exp:idm}

As an agent progresses through the environment, new visual dynamics emerge, and the IDM must correctly label these new dynamics for policy bootstrapping to be effective.
To measure how well the IDM captures evolving dynamics, we take the checkpoint from each bootstrap of an \method{} playthrough of \Pokemon{} Emerald, and evaluate it on a test set of trajectories from a separate ASH playthrough. We report the per-bootstrap IDM accuracy in \Cref{fig:idm} (right).
We find that the IDM becomes better at labeling environment dynamics across bootstraps. The initialization IDM checkpoint (no bootstrapping baseline) has just 60\% accuracy on our test set, whereas our final IDM checkpoint achieves 81\%.
We limit our analysis to the four (up, down, left, right) movement action classes since some actions like the \texttt{A} and \texttt{B} buttons are often visually indistinguishable.

We find that IDM bootstrapping is essential for progression. In \Pokemon{} Emerald, training the VPT and BC baselines using our curated dataset and the no bootstrapping baseline IDM yielded 0\% completion rates on milestones 3 and above.

\begin{wraptable}{R}{0.49\textwidth}
\vspace{-3ex}
\centering
\small
\begin{tabular}{lcccc}
\toprule
& \method{} & MSM & All2All & Rand \\
\midrule
Binary Rel. $\uparrow$ & \textbf{94\%} & \underline{69\%} & 3\% & 8\% \\
Ave. Rank $\downarrow$ & \textbf{1.35} & \underline{1.81} & 3.63 & 3.29 \\
\bottomrule
\end{tabular}
\caption{Annotator judgments of \retmet{} and baselines: binary relevance (yes/no) and average rank across methods (lower is better).}
\label{tab:retrieval_judging_summary}
\vspace{-2ex}
\end{wraptable}

\subsection{Retrieval Quality}
\label{exp:ret}

For bootstrapping to be effective, it is crucial that \retmet{} creates a \emph{relevant and low-noise} training dataset $\DR$ from our corpus of YouTube data, $\DI$.
To measure the effectiveness of \retmet{}, we conduct a blind evaluation. For each query trajectory, an annotator sees the videos retrieved by each method and performs two tasks. First, they label each retrieved video as relevant training data (yes/no). Second, given one video per method, they rank the four videos using the following criteria, in priority order:
\begin{enumerate}[itemsep=1pt, parsep=0pt, topsep=2pt]
    \item Does the retrieved video correspond to the same part of the game as the trajectory?
    \item Is the retrieved video visually similar to the trajectory?
    \item Does the retrieved video contain future milestones?
\end{enumerate}
Using the above criteria, the annotators rank the videos from 1 (most relevant) to 4 (least relevant).
For each method considered, we use {10} trajectories sampled from bootstrapping runs of \Pokemon{} Emerald (\Cref{exp:main}).
We judge the top {20} videos per trajectory, for a total of 800 videos judged.

We compare \textbf{\retmet{}} with three baselines. \textbf{Random} selects videos uniformly from the entire dataset. \textbf{All-to-All} ranks videos by the average similarity between all trajectory and internet video embeddings. \textbf{Max Single Match} replaces the sliding window score of \retmet{} with the maximum score of any given trajectory embedding with an internet video embedding.
We report our findings in \Cref{tab:retrieval_judging_summary}.
As \pokemon{} is a long-horizon game, we find a randomly retrieved video has only an 8\% chance of being relevant training data, highlighting the challenges of naively training a policy over $\DI$.
By contrast, \method{} retrieves relevant videos 94\% of the time, achieving an average rank of 1.35 between methods. We find that Max Single Match occasionally makes spurious matches, for example matching a black screen between videos, and All-to-All matching seems to lose all signal, consistently matching to the same internet video, irrespective of the query.

\subsection{Long-term Memory Analysis}
\label{exp:clustering}
\begin{figure}[h]
  \centering
  \includegraphics[width=\linewidth]{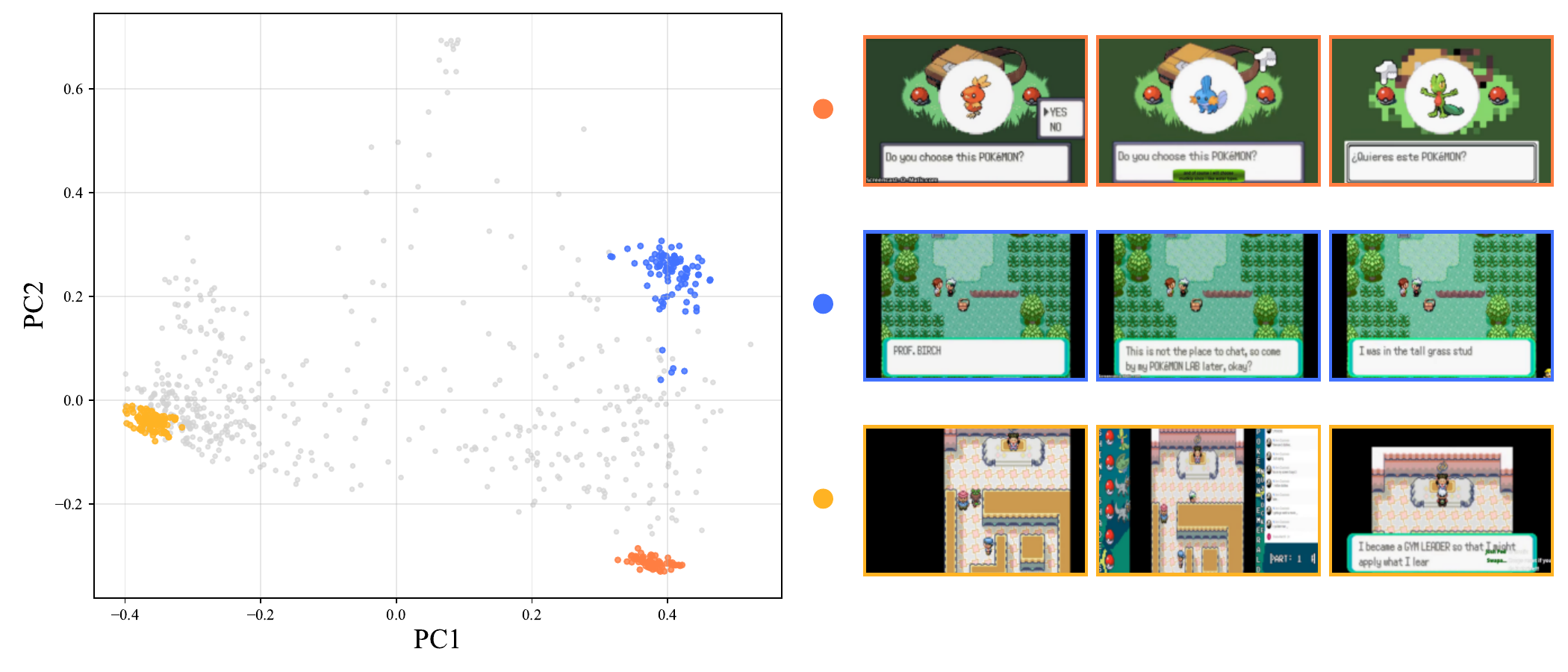}
  \vspace{-0.6cm}
\caption{
  Visualization of 3 HDBSCAN~\citep{hbscan} clusters, as well as 500 uniformly sampled outlier points reduced to 2 dimensions via principal component analysis. Each of these clusters represents a key moment in \pokemon{} Emerald; choosing a starter \pokemon{} (orange), saving the professor (blue) and challenging a gym leader (yellow). Grey dots are outliers that are not assigned to a cluster. 
  }
\label{fig:cluster_analysis}
\vspace{-0.3cm}
\end{figure}

The effectiveness of \method{}'s long-term memory depends on the quality of the clusters produced by HDBSCAN~\citep{hbscan}. A cluster should be composed of data from a single \emph{key moment} --- an observation that can be used to contextualize a player's actions within the broader trajectory. Examples of key moments include reaching a milestone, selecting your starter \pokemon{}, or battling a particular enemy.

To evaluate how well HDBSCAN~\citep{hbscan} clusters these key moments, we evaluated 400 clusters sampled uniformly at random from a pool of 2,200 clusters generated by \method{} across all bootstraps during an eight-hour run in \pokemon{} Emerald. For each cluster, annotators were shown five of its constituent internet video frames and answered two yes/no questions:
\begin{enumerate}[itemsep=1pt, parsep=0pt, topsep=2pt]
    \item Are all frames \emph{key moments}?
    \item Do all frames correspond to the \emph{same} key moment?
\end{enumerate}

\begin{wraptable}{R}{0.5\textwidth}
\vspace{+0ex}
\centering
\small
\begin{tabular}{lc}
\toprule
Long-term memory contents & Avg.\ milestones \\
\midrule
Random trajectory frames & $6.00$ \\
Key moments (ours)       & $\mathbf{7.75}$ \\
\bottomrule
\end{tabular}
\caption{\label{tab:random_memory}Average milestones achieved for key moments versus random frames as the contents of long-term memory, in \Pokemon{} Emerald. Each value is an average over 4 runs of 4 hours.}
\vspace{-2ex}
\end{wraptable}
\noindent
If the answer to both questions is yes, we consider the cluster to be a key moment. For comparison, we also evaluate 400 randomly sampled frames from the agent's trajectory. We find that 48\% of clusters identified by HDBSCAN~\citep{hbscan} correspond to key moments, compared to just 0.5\% of randomly sampled frames.

To validate the usefulness of these HDBSCAN identified key moments, we replace the $\longlen$ key moments with $\longlen$ frames sampled uniformly at random from the agent's trajectory during inference. Since we cannot perform dynamic bootstrapping with randomly sampled long memory, we compare against Bootstrapped Agent + Long Memory from \Cref{sec:ablation} averaging over 4 trajectories of 4 hours each in \Pokemon{} Emerald. We report results in \Cref{tab:random_memory}.

We find that replacing key moments with random frames reduces average achieved milestones by $1.75$ over the same evaluation period. The benefit of long-term memory therefore comes from \emph{which} observations \method{} keeps, indicating the usefulness of HDBSCAN identified key moments.

\subsection{Offline Replay}
\label{exp:forget}

\begin{wrapfigure}[15]{r}{0.5\linewidth}
\vspace{-2ex}
\centering
\includegraphics[width=\linewidth]{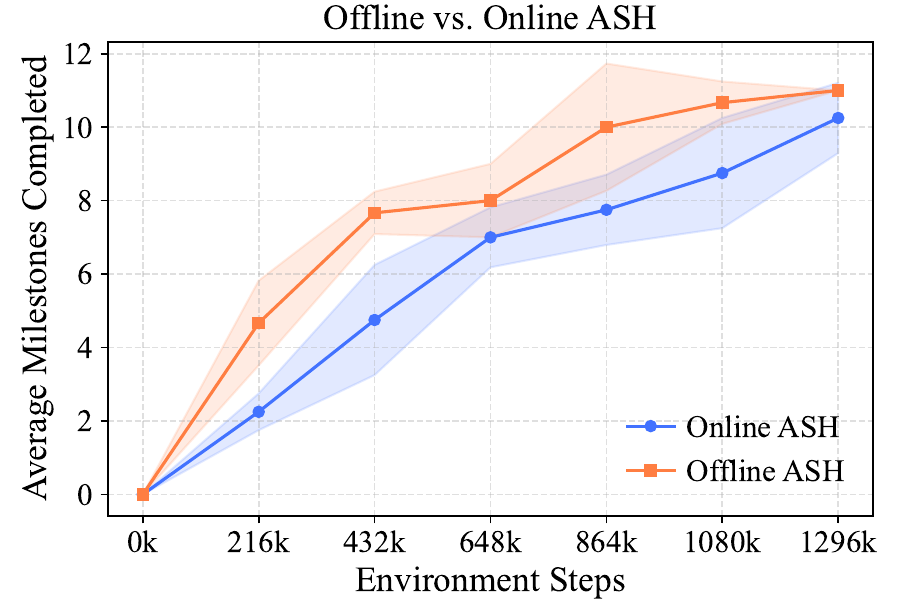}
\vspace{-4ex}
\caption{Offline replay of the final \method{} checkpoint vs.\ the original online run.}
\label{fig:forget}
\vspace{-2ex}
\end{wrapfigure}
\emph{Catastrophic forgetting} is a phenomenon in lifelong learning where an agent will forget previously known skills and knowledge when its policy is updated~\citep{10.1109/TPAMI.2024.3498346}.
The result is an agent that can progress through the latter stages of an environment but can no longer accomplish early milestones.
We examine whether \method{}'s final policy is able to play through the environment from the beginning without bootstrapping, or if it exhibits catastrophic forgetting.

We compare the final \method{} checkpoint (offline ASH) with the online run that created it (online ASH). Both methods are run for 1.3 million environment steps (6 hours) in \Pokemon{} Emerald~(\Cref{fig:forget}). Results are averaged over 4 trajectories and the shaded regions show the standard deviation. We find that \method{} does not exhibit catastrophic forgetting.
As the offline \method{} checkpoint has already learned (and remembers) the environment, it progresses through milestones faster than the online learning agent. 
This is analogous to a player replaying the environment, who, due to accumulated expertise, is faster than the beginner.

\vspace{-2ex}
\section{Limitations and Conclusion}
\label{sec:conclusion}
\vspace{-1ex}

We introduced \method{}, an agent that self-improves by retrieving and learning from relevant internet video, without reward engineering or human annotation. Across \Pokemon{} Emerald and The Legend of Zelda, \method{}
continues to adapt beyond the point where behavioral cloning, retrieval-augmented, and foundation-model baselines stall, developing new skills such as battling, catching \pokemon{}, and using a sword and shield. These results show that self-improving agents are a scalable path toward open-ended, long-horizon learning.

Although \method{} improves over fixed policies in long-horizon settings, it depends on the availability of useful internet video and may be harder to apply in environments where demonstrations are scarce. It also relies on pseudo-actions from an Inverse Dynamics Model, which will struggle to label actions that are difficult to infer from pixels. Furthermore, as \method{} was evaluated in 2D environments with discrete action spaces, it remains unclear how well \method{} generalizes to 3D environments, or to those with continuous action spaces.
%
%
Finally, since our approach is not specific to \pokemon{} or Zelda, we hope \method{} can inspire future work in creating long-horizon agents in other domains and environments.

\begin{ack}
This research was supported by the New Beginnings program of the National Research Council of Canada (INBR6-001156), as well as the Natural Sciences and Engineering Research Council of Canada (NSERC) through Victor Zhong's Discovery Grant, Research Tools and Instruments grant, and by the Canada Foundation for Innovation (CFI) John R. Evans Leaders Fund with matching support from the Ontario Research Fund.
Furthermore, this research was enabled by computing resources provided by Compute Ontario and the Digital Research Alliance of Canada.
\end{ack}

{\small
\bibliographystyle{unsrtnat}
\bibliography{references}

@misc{pleines2025pokemon,
    title={Pokemon Red via Reinforcement Learning},
    author={Marco Pleines and Daniel Addis and David Rubinstein and Frank Zimmer and Mike Preuss and Peter Whidden},
    year={2025},
    eprint={2502.19920},
    archivePrefix={arXiv},
    primaryClass={cs.LG}
}

@misc{hambro2023dungeonsdatalargescalenethack,
      title={Dungeons and Data: A Large-Scale NetHack Dataset}, 
      author={Eric Hambro and Roberta Raileanu and Danielle Rothermel and Vegard Mella and Tim Rocktäschel and Heinrich Küttler and Naila Murray},
      year={2023},
      eprint={2211.00539},
      archivePrefix={arXiv},
      primaryClass={cs.LG},
      url={https://arxiv.org/abs/2211.00539}, 
}

@inproceedings{karten2025pokeagent,
  title        = {The PokeAgent Challenge: Competitive and Long-Context Learning at Scale},
  author       = {Karten, Seth and Grigsby, Jake and Milani, Stephanie and Vodrahalli, Kiran
                  and Zhang, Amy and Fang, Fei and Zhu, Yuke and Jin, Chi},
  booktitle    = {NeurIPS Competition Track},
  year         = {2025},
  month        = apr,
}

@misc{hafner2025trainingagentsinsidescalable,
      title={Training Agents Inside of Scalable World Models}, 
      author={Danijar Hafner and Wilson Yan and Timothy Lillicrap},
      year={2025},
      eprint={2509.24527},
      archivePrefix={arXiv},
      primaryClass={cs.AI},
      url={https://arxiv.org/abs/2509.24527}, 
}

@misc{baker2022videopretrainingvptlearning,
      title={Video PreTraining (VPT): Learning to Act by Watching Unlabeled Online Videos}, 
      author={Bowen Baker and Ilge Akkaya and Peter Zhokhov and Joost Huizinga and Jie Tang and Adrien Ecoffet and Brandon Houghton and Raul Sampedro and Jeff Clune},
      year={2022},
      eprint={2206.11795},
      archivePrefix={arXiv},
      primaryClass={cs.LG},
      url={https://arxiv.org/abs/2206.11795}, 
}

@misc{zhai2023sigmoidlosslanguageimage,
      title={Sigmoid Loss for Language Image Pre-Training}, 
      author={Xiaohua Zhai and Basil Mustafa and Alexander Kolesnikov and Lucas Beyer},
      year={2023},
      eprint={2303.15343},
      archivePrefix={arXiv},
      primaryClass={cs.CV},
      url={https://arxiv.org/abs/2303.15343}, 
}

@misc{oquab2024dinov2learningrobustvisual,
      title={DINOv2: Learning Robust Visual Features without Supervision}, 
      author={Maxime Oquab and Timothée Darcet and Théo Moutakanni and Huy Vo and Marc Szafraniec and Vasil Khalidov and Pierre Fernandez and Daniel Haziza and Francisco Massa and Alaaeldin El-Nouby and Mahmoud Assran and Nicolas Ballas and Wojciech Galuba and Russell Howes and Po-Yao Huang and Shang-Wen Li and Ishan Misra and Michael Rabbat and Vasu Sharma and Gabriel Synnaeve and Hu Xu and Hervé Jegou and Julien Mairal and Patrick Labatut and Armand Joulin and Piotr Bojanowski},
      year={2024},
      eprint={2304.07193},
      archivePrefix={arXiv},
      primaryClass={cs.CV},
      url={https://arxiv.org/abs/2304.07193}, 
}

@misc{chen2021decisiontransformerreinforcementlearning,
      title={Decision Transformer: Reinforcement Learning via Sequence Modeling}, 
      author={Lili Chen and Kevin Lu and Aravind Rajeswaran and Kimin Lee and Aditya Grover and Michael Laskin and Pieter Abbeel and Aravind Srinivas and Igor Mordatch},
      year={2021},
      eprint={2106.01345},
      archivePrefix={arXiv},
      primaryClass={cs.LG},
      url={https://arxiv.org/abs/2106.01345}, 
}

@misc{mnih2013playingatarideepreinforcement,
      title={Playing Atari with Deep Reinforcement Learning}, 
      author={Volodymyr Mnih and Koray Kavukcuoglu and David Silver and Alex Graves and Ioannis Antonoglou and Daan Wierstra and Martin Riedmiller},
      year={2013},
      eprint={1312.5602},
      archivePrefix={arXiv},
      primaryClass={cs.LG},
      url={https://arxiv.org/abs/1312.5602}, 
}

@misc{openai2019dota2largescale,
      title={Dota 2 with Large Scale Deep Reinforcement Learning}, 
      author={OpenAI and : and Christopher Berner and Greg Brockman and Brooke Chan and Vicki Cheung and Przemysław Dębiak and Christy Dennison and David Farhi and Quirin Fischer and Shariq Hashme and Chris Hesse and Rafal Józefowicz and Scott Gray and Catherine Olsson and Jakub Pachocki and Michael Petrov and Henrique P. d. O. Pinto and Jonathan Raiman and Tim Salimans and Jeremy Schlatter and Jonas Schneider and Szymon Sidor and Ilya Sutskever and Jie Tang and Filip Wolski and Susan Zhang},
      year={2019},
      eprint={1912.06680},
      archivePrefix={arXiv},
      primaryClass={cs.LG},
      url={https://arxiv.org/abs/1912.06680}, 
}

@misc{Bulbapedia_2025,
title={Walkthrough:pokémon emerald},
url={https://bulbapedia.bulbagarden.net/wiki/Walkthrough:Pok%C3%A9mon_Emerald},
publisher={Bulbapedia},
author={Bulbapedia},
year={2025}, month={Jun}}

@misc{yang2025qwen3technicalreport,
      title={Qwen3 Technical Report}, 
      author={An Yang and Anfeng Li and Baosong Yang and Beichen Zhang and Binyuan Hui and Bo Zheng and Bowen Yu and Chang Gao and Chengen Huang and Chenxu Lv and Chujie Zheng and Dayiheng Liu and Fan Zhou and Fei Huang and Feng Hu and Hao Ge and Haoran Wei and Huan Lin and Jialong Tang and Jian Yang and Jianhong Tu and Jianwei Zhang and Jianxin Yang and Jiaxi Yang and Jing Zhou and Jingren Zhou and Junyang Lin and Kai Dang and Keqin Bao and Kexin Yang and Le Yu and Lianghao Deng and Mei Li and Mingfeng Xue and Mingze Li and Pei Zhang and Peng Wang and Qin Zhu and Rui Men and Ruize Gao and Shixuan Liu and Shuang Luo and Tianhao Li and Tianyi Tang and Wenbiao Yin and Xingzhang Ren and Xinyu Wang and Xinyu Zhang and Xuancheng Ren and Yang Fan and Yang Su and Yichang Zhang and Yinger Zhang and Yu Wan and Yuqiong Liu and Zekun Wang and Zeyu Cui and Zhenru Zhang and Zhipeng Zhou and Zihan Qiu},
      year={2025},
      eprint={2505.09388},
      archivePrefix={arXiv},
      primaryClass={cs.CL},
      url={https://arxiv.org/abs/2505.09388}, 
}

@misc{loshchilov2019decoupledweightdecayregularization,
      title={Decoupled Weight Decay Regularization}, 
      author={Ilya Loshchilov and Frank Hutter},
      year={2019},
      eprint={1711.05101},
      archivePrefix={arXiv},
      primaryClass={cs.LG},
      url={https://arxiv.org/abs/1711.05101}, 
}

@InProceedings{hbscan,
author="Campello, Ricardo J. G. B.
and Moulavi, Davoud
and Sander, Joerg",
editor="Pei, Jian
and Tseng, Vincent S.
and Cao, Longbing
and Motoda, Hiroshi
and Xu, Guandong",
title="Density-Based Clustering Based on Hierarchical Density Estimates",
booktitle="Advances in Knowledge Discovery and Data Mining",
year="2013",
publisher="Springer Berlin Heidelberg",
address="Berlin, Heidelberg",
pages="160--172",
isbn="978-3-642-37456-2"
}

@inproceedings{ijcai2018p687,
  title     = {Behavioral Cloning from Observation},
  author    = {Faraz Torabi and Garrett Warnell and Peter Stone},
  booktitle = {Proceedings of the Twenty-Seventh International Joint Conference on
               Artificial Intelligence, {IJCAI-18}},
  publisher = {International Joint Conferences on Artificial Intelligence Organization},
  pages     = {4950--4957},
  year      = {2018},
  month     = {7},
  doi       = {10.24963/ijcai.2018/687},
  url       = {https://doi.org/10.24963/ijcai.2018/687},
}

@inproceedings{li2025optimus2,
    title={Optimus-2: Multimodal Minecraft Agent with Goal-Observation-Action Conditioned Policy},
    author={Li, Zaijing and Xie, Yuquan and Shao, Rui and Chen, Gongwei and Jiang, Dongmei and Nie, Liqiang},
    booktitle={2025 IEEE/CVF Conference on Computer Vision and Pattern Recognition (CVPR)},
    year={2025},
    organization={IEEE}
}

@InProceedings{pmlr-v9-ross10a,
  title = 	 {Efficient Reductions for Imitation Learning},
  author = 	 {Ross, Stephane and Bagnell, Drew},
  booktitle = 	 {Proceedings of the Thirteenth International Conference on Artificial Intelligence and Statistics},
  pages = 	 {661--668},
  year = 	 {2010},
  editor = 	 {Teh, Yee Whye and Titterington, Mike},
  volume = 	 {9},
  series = 	 {Proceedings of Machine Learning Research},
  address = 	 {Chia Laguna Resort, Sardinia, Italy},
  month = 	 {13--15 May},
  publisher =    {PMLR},
  url = 	 {https://proceedings.mlr.press/v9/ross10a.html}
}

@misc{qwen35blog,
    title = {Qwen3.5: Accelerating Productivity with Native Multimodal Agents},
    url = {https://qwen.ai/blog?id=qwen3.5},
    author = {{Qwen Team}},
    month = {February},
    year = {2026}
}

@inproceedings{fan2022minedojo,
  title     = {MineDojo: Building Open-Ended Embodied Agents with Internet-Scale Knowledge},
  author    = {Linxi Fan and Guanzhi Wang and Yunfan Jiang and Ajay Mandlekar and Yuncong Yang and Haoyi Zhu and Andrew Tang and De-An Huang and Yuke Zhu and Anima Anandkumar},
  booktitle = {Thirty-sixth Conference on Neural Information Processing Systems Datasets and Benchmarks Track},
  year      = {2022},
  url       = {https://openreview.net/forum?id=rc8o_j8I8PX}
}

@misc{hu2024generalpurposerobotsfoundationmodels,
      title={Toward General-Purpose Robots via Foundation Models: A Survey and Meta-Analysis}, 
      author={Yafei Hu and Quanting Xie and Vidhi Jain and Jonathan Francis and Jay Patrikar and Nikhil Keetha and Seungchan Kim and Yaqi Xie and Tianyi Zhang and Hao-Shu Fang and Shibo Zhao and Shayegan Omidshafiei and Dong-Ki Kim and Ali-akbar Agha-mohammadi and Katia Sycara and Matthew Johnson-Roberson and Dhruv Batra and Xiaolong Wang and Sebastian Scherer and Chen Wang and Zsolt Kira and Fei Xia and Yonatan Bisk},
      year={2024},
      eprint={2312.08782},
      archivePrefix={arXiv},
      primaryClass={cs.RO},
      url={https://arxiv.org/abs/2312.08782}, 
}

@misc{huang2026rethinkingmemorymechanismsfoundation,
      title={Rethinking Memory Mechanisms of Foundation Agents in the Second Half: A Survey}, 
      author={Wei-Chieh Huang and Weizhi Zhang and Yueqing Liang and Yuanchen Bei and Yankai Chen and Tao Feng and Xinyu Pan and Zhen Tan and Yu Wang and Tianxin Wei and Shanglin Wu and Ruiyao Xu and Liangwei Yang and Rui Yang and Wooseong Yang and Chin-Yuan Yeh and Hanrong Zhang and Haozhen Zhang and Siqi Zhu and Henry Peng Zou and Wanjia Zhao and Song Wang and Wujiang Xu and Zixuan Ke and Zheng Hui and Dawei Li and Yaozu Wu and Langzhou He and Chen Wang and Xiongxiao Xu and Baixiang Huang and Juntao Tan and Shelby Heinecke and Huan Wang and Caiming Xiong and Ahmed A. Metwally and Jun Yan and Chen-Yu Lee and Hanqing Zeng and Yinglong Xia and Xiaokai Wei and Ali Payani and Yu Wang and Haitong Ma and Wenya Wang and Chenguang Wang and Yu Zhang and Xin Wang and Yongfeng Zhang and Jiaxuan You and Hanghang Tong and Xiao Luo and Xue Liu and Yizhou Sun and Wei Wang and Julian McAuley and James Zou and Jiawei Han and Philip S. Yu and Kai Shu},
      year={2026},
      eprint={2602.06052},
      archivePrefix={arXiv},
      primaryClass={cs.CL},
      url={https://arxiv.org/abs/2602.06052}, 
}

@misc{wang2023voyageropenendedembodiedagent,
      title={Voyager: An Open-Ended Embodied Agent with Large Language Models}, 
      author={Guanzhi Wang and Yuqi Xie and Yunfan Jiang and Ajay Mandlekar and Chaowei Xiao and Yuke Zhu and Linxi Fan and Anima Anandkumar},
      year={2023},
      eprint={2305.16291},
      archivePrefix={arXiv},
      primaryClass={cs.AI},
      url={https://arxiv.org/abs/2305.16291}, 
}

@misc{minishcap,
	author = {{Zelda Dungeon}},
	title = {The Minish Cap Walkthrough},
	howpublished = {\url{https://www.zeldadungeon.net/the-minish-cap-walkthrough/}},
	year = {2026},
	note = {[Accessed 30-04-2026]},
}

@misc{torabi2018behavioralcloningobservation,
      title={Behavioral Cloning from Observation}, 
      author={Faraz Torabi and Garrett Warnell and Peter Stone},
      year={2018},
      eprint={1805.01954},
      archivePrefix={arXiv},
      primaryClass={cs.AI},
      url={https://arxiv.org/abs/1805.01954}, 
}

@InProceedings{pmlr-v97-edwards19a,
  title = 	 {Imitating Latent Policies from Observation},
  author =       {Edwards, Ashley and Sahni, Himanshu and Schroecker, Yannick and Isbell, Charles},
  booktitle = 	 {Proceedings of the 36th International Conference on Machine Learning},
  pages = 	 {1755--1763},
  year = 	 {2019},
  editor = 	 {Chaudhuri, Kamalika and Salakhutdinov, Ruslan},
  volume = 	 {97},
  series = 	 {Proceedings of Machine Learning Research},
  month = 	 {09--15 Jun},
  publisher =    {PMLR},
  url = 	 {https://proceedings.mlr.press/v97/edwards19a.html}
}

@inproceedings{lapo,
  title={Learning to Act without Actions},
  author={Schmidt, Dominik and Jiang, Minqi},
  booktitle={The Twelfth International Conference on Learning Representations (ICLR)},
  year={2024}
}

@InProceedings{pmlr-v235-bruce24a,
  title = 	 {Genie: Generative Interactive Environments},
  author =       {Bruce, Jake and Dennis, Michael D and Edwards, Ashley and Parker-Holder, Jack and Shi, Yuge and Hughes, Edward and Lai, Matthew and Mavalankar, Aditi and Steigerwald, Richie and Apps, Chris and Aytar, Yusuf and Bechtle, Sarah Maria Elisabeth and Behbahani, Feryal and Chan, Stephanie C.Y. and Heess, Nicolas and Gonzalez, Lucy and Osindero, Simon and Ozair, Sherjil and Reed, Scott and Zhang, Jingwei and Zolna, Konrad and Clune, Jeff and Freitas, Nando De and Singh, Satinder and Rockt\"{a}schel, Tim},
  booktitle = 	 {Proceedings of the 41st International Conference on Machine Learning},
  pages = 	 {4603--4623},
  year = 	 {2024},
  editor = 	 {Salakhutdinov, Ruslan and Kolter, Zico and Heller, Katherine and Weller, Adrian and Oliver, Nuria and Scarlett, Jonathan and Berkenkamp, Felix},
  volume = 	 {235},
  series = 	 {Proceedings of Machine Learning Research},
  month = 	 {21--27 Jul},
  publisher =    {PMLR},
  url = 	 {https://proceedings.mlr.press/v235/bruce24a.html}
}

@inproceedings{lewis2020rag,
author = {Lewis, Patrick and Perez, Ethan and Piktus, Aleksandra and Petroni, Fabio and Karpukhin, Vladimir and Goyal, Naman and K\"{u}ttler, Heinrich and Lewis, Mike and Yih, Wen-tau and Rockt\"{a}schel, Tim and Riedel, Sebastian and Kiela, Douwe},
title = {Retrieval-augmented generation for knowledge-intensive NLP tasks},
year = {2020},
isbn = {9781713829546},
publisher = {Curran Associates Inc.},
address = {Red Hook, NY, USA},
booktitle = {Proceedings of the 34th International Conference on Neural Information Processing Systems},
articleno = {793},
numpages = {16},
location = {Vancouver, BC, Canada},
series = {NIPS '20}
}

@article{yao2022react,
  title={ReAct: Synergizing Reasoning and Acting in Language Models},
  author={Yao, Shunyu and Zhao, Jeffrey and Yu, Dian and Du, Nan and Shafran, Izhak and Narasimhan, Karthik and Cao, Yuan},
  journal={arXiv preprint arXiv:2210.03629},
  year={2022}
}

@inproceedings{
shinn2023reflexion,
title={Reflexion: language agents with verbal reinforcement learning},
author={Noah Shinn and Federico Cassano and Ashwin Gopinath and Karthik R Narasimhan and Shunyu Yao},
booktitle={Thirty-seventh Conference on Neural Information Processing Systems},
year={2023},
url={https://openreview.net/forum?id=vAElhFcKW6}
}

@inproceedings{
humphreys2022largescale,
title={Large-Scale Retrieval for Reinforcement Learning},
author={Peter Conway Humphreys and Arthur Guez and Olivier Tieleman and Laurent Sifre and Theophane Weber and Timothy P Lillicrap},
booktitle={Advances in Neural Information Processing Systems},
editor={Alice H. Oh and Alekh Agarwal and Danielle Belgrave and Kyunghyun Cho},
year={2022},
url={https://openreview.net/forum?id=Ya9lATuQ3gg}
}

@InProceedings{pmlr-v162-goyal22a,
  title = 	 {Retrieval-Augmented Reinforcement Learning},
  author =       {Goyal, Anirudh and Friesen, Abram and Banino, Andrea and Weber, Theophane and Ke, Nan Rosemary and Badia, Adri{\`a} Puigdom{\`e}nech and Guez, Arthur and Mirza, Mehdi and Humphreys, Peter C and Konyushova, Ksenia and Valko, Michal and Osindero, Simon and Lillicrap, Timothy and Heess, Nicolas and Blundell, Charles},
  booktitle = 	 {Proceedings of the 39th International Conference on Machine Learning},
  pages = 	 {7740--7765},
  year = 	 {2022},
  editor = 	 {Chaudhuri, Kamalika and Jegelka, Stefanie and Song, Le and Szepesvari, Csaba and Niu, Gang and Sabato, Sivan},
  volume = 	 {162},
  series = 	 {Proceedings of Machine Learning Research},
  month = 	 {17--23 Jul},
  publisher =    {PMLR},
  url = 	 {https://proceedings.mlr.press/v162/goyal22a.html}
}

@inproceedings{
zhang2023large,
title={Large Language Models Are Semi-Parametric Reinforcement Learning Agents},
author={Danyang Zhang and Lu Chen and Situo Zhang and Hongshen Xu and Zihan Zhao and Kai Yu},
booktitle={Thirty-seventh Conference on Neural Information Processing Systems},
year={2023},
url={https://openreview.net/forum?id=ZcJa1R6j3v}
}

@inproceedings{
zelikman2022star,
title={{ST}aR: Bootstrapping Reasoning With Reasoning},
author={Eric Zelikman and Yuhuai Wu and Jesse Mu and Noah Goodman},
booktitle={Advances in Neural Information Processing Systems},
editor={Alice H. Oh and Alekh Agarwal and Danielle Belgrave and Kyunghyun Cho},
year={2022},
url={https://openreview.net/forum?id=_3ELRdg2sgI}
}

@misc{gulcehre2023reinforcedselftrainingrestlanguage,
      title={Reinforced Self-Training (ReST) for Language Modeling}, 
      author={Caglar Gulcehre and Tom Le Paine and Srivatsan Srinivasan and Ksenia Konyushkova and Lotte Weerts and Abhishek Sharma and Aditya Siddhant and Alex Ahern and Miaosen Wang and Chenjie Gu and Wolfgang Macherey and Arnaud Doucet and Orhan Firat and Nando de Freitas},
      year={2023},
      eprint={2308.08998},
      archivePrefix={arXiv},
      primaryClass={cs.CL},
      url={https://arxiv.org/abs/2308.08998}, 
}

@inproceedings{10.5555/3295222.3295288,
author = {Anthony, Thomas and Tian, Zheng and Barber, David},
title = {Thinking fast and slow with deep learning and tree search},
year = {2017},
isbn = {9781510860964},
publisher = {Curran Associates Inc.},
address = {Red Hook, NY, USA},
booktitle = {Proceedings of the 31st International Conference on Neural Information Processing Systems},
pages = {5366–5376},
numpages = {11},
location = {Long Beach, California, USA},
series = {NIPS'17}
}

@misc{silver2017masteringchessshogiselfplay,
      title={Mastering Chess and Shogi by Self-Play with a General Reinforcement Learning Algorithm}, 
      author={David Silver and Thomas Hubert and Julian Schrittwieser and Ioannis Antonoglou and Matthew Lai and Arthur Guez and Marc Lanctot and Laurent Sifre and Dharshan Kumaran and Thore Graepel and Timothy Lillicrap and Karen Simonyan and Demis Hassabis},
      year={2017},
      eprint={1712.01815},
      archivePrefix={arXiv},
      primaryClass={cs.AI},
      url={https://arxiv.org/abs/1712.01815}, 
}

@misc{gdm2024autort,
      title={AutoRT: Embodied Foundation Models for Large Scale Orchestration of Robotic Agents}, 
      author={Michael Ahn and Debidatta Dwibedi and Chelsea Finn and Montse Gonzalez Arenas and Keerthana Gopalakrishnan and Karol Hausman and Brian Ichter and Alex Irpan and Nikhil Joshi and Ryan Julian and Sean Kirmani and Isabel Leal and Edward Lee and Sergey Levine and Yao Lu and Isabel Leal and Sharath Maddineni and Kanishka Rao and Dorsa Sadigh and Pannag Sanketi and Pierre Sermanet and Quan Vuong and Stefan Welker and Fei Xia and Ted Xiao and Peng Xu and Steve Xu and Zhuo Xu},
      year={2024},
      eprint={2401.12963},
      archivePrefix={arXiv},
      primaryClass={cs.RO}
}

@inproceedings{
wu2022memorizing,
title={Memorizing Transformers},
author={Yuhuai Wu and Markus Norman Rabe and DeLesley Hutchins and Christian Szegedy},
booktitle={International Conference on Learning Representations},
year={2022},
url={https://openreview.net/forum?id=TrjbxzRcnf-}
}

@inproceedings{
bulatov2022recurrent,
title={Recurrent Memory Transformer},
author={Aydar Bulatov and Yuri Kuratov and Mikhail Burtsev},
booktitle={Advances in Neural Information Processing Systems},
editor={Alice H. Oh and Alekh Agarwal and Danielle Belgrave and Kyunghyun Cho},
year={2022},
url={https://openreview.net/forum?id=Uynr3iPhksa}
}

@InProceedings{pmlr-v162-borgeaud22a,
  title = 	 {Improving Language Models by Retrieving from Trillions of Tokens},
  author =       {Borgeaud, Sebastian and Mensch, Arthur and Hoffmann, Jordan and Cai, Trevor and Rutherford, Eliza and Millican, Katie and Van Den Driessche, George Bm and Lespiau, Jean-Baptiste and Damoc, Bogdan and Clark, Aidan and De Las Casas, Diego and Guy, Aurelia and Menick, Jacob and Ring, Roman and Hennigan, Tom and Huang, Saffron and Maggiore, Loren and Jones, Chris and Cassirer, Albin and Brock, Andy and Paganini, Michela and Irving, Geoffrey and Vinyals, Oriol and Osindero, Simon and Simonyan, Karen and Rae, Jack and Elsen, Erich and Sifre, Laurent},
  booktitle = 	 {Proceedings of the 39th International Conference on Machine Learning},
  pages = 	 {2206--2240},
  year = 	 {2022},
  editor = 	 {Chaudhuri, Kamalika and Jegelka, Stefanie and Song, Le and Szepesvari, Csaba and Niu, Gang and Sabato, Sivan},
  volume = 	 {162},
  series = 	 {Proceedings of Machine Learning Research},
  month = 	 {17--23 Jul},
  publisher =    {PMLR},
  url = 	 {https://proceedings.mlr.press/v162/borgeaud22a.html}
}

@article{Vinyals2019GrandmasterLI,
  title={Grandmaster level in StarCraft II using multi-agent reinforcement learning},
  author={Oriol Vinyals and Igor Babuschkin and Wojciech M. Czarnecki and Micha{\"e}l Mathieu and Andrew Joseph Dudzik and Junyoung Chung and David Choi and Richard Powell and Timo Ewalds and Petko Georgiev and Junhyuk Oh and Dan Horgan and Manuel Kroiss and Ivo Danihelka and Aja Huang and L. Sifre and Trevor Cai and John P. Agapiou and Max Jaderberg and Alexander Sasha Vezhnevets and R{\'e}mi Leblond and Tobias Pohlen and Valentin Dalibard and David Budden and Yury Sulsky and James Molloy and Tom Le Paine and Caglar Gulcehre and Ziyun Wang and Tobias Pfaff and Yuhuai Wu and Roman Ring and Dani Yogatama and Dario W{\"u}nsch and Katrina McKinney and Oliver Smith and Tom Schaul and Timothy P. Lillicrap and Koray Kavukcuoglu and Demis Hassabis and Chris Apps and David Silver},
  journal={Nature},
  year={2019},
  volume={575},
  pages={350 - 354},
  url={https://api.semanticscholar.org/CorpusID:204972004}
}

@misc{twitchClaudePlaysPokemonTwitch,
	author = {ClaudePlaysPokemon},
	title = {{C}laude{P}lays{P}okemon},
	howpublished = {\url{https://www.twitch.tv/claudeplayspokemon}},
	year = {2026},
	note = {[Accessed 01-05-2026]},
}

@misc{simateam2025sima2generalistembodied,
      title={SIMA 2: A Generalist Embodied Agent for Virtual Worlds}, 
      author={SIMA team and Adrian Bolton and Alexander Lerchner and Alexandra Cordell and Alexandre Moufarek and Andrew Bolt and Andrew Lampinen and Anna Mitenkova and Arne Olav Hallingstad and Bojan Vujatovic and Bonnie Li and Cong Lu and Daan Wierstra and Daniel P. Sawyer and Daniel Slater and David Reichert and Davide Vercelli and Demis Hassabis and Drew A. Hudson and Duncan Williams and Ed Hirst and Fabio Pardo and Felix Hill and Frederic Besse and Hannah Openshaw and Harris Chan and Hubert Soyer and Jane X. Wang and Jeff Clune and John Agapiou and John Reid and Joseph Marino and Junkyung Kim and Karol Gregor and Kaustubh Sridhar and Kay McKinney and Laura Kampis and Lei M. Zhang and Loic Matthey and Luyu Wang and Maria Abi Raad and Maria Loks-Thompson and Martin Engelcke and Matija Kecman and Matthew Jackson and Maxime Gazeau and Ollie Purkiss and Oscar Knagg and Peter Stys and Piermaria Mendolicchio and Raia Hadsell and Rosemary Ke and Ryan Faulkner and Sarah Chakera and Satinder Singh Baveja and Shane Legg and Sheleem Kashem and Tayfun Terzi and Thomas Keck and Tim Harley and Tim Scholtes and Tyson Roberts and Volodymyr Mnih and Yulan Liu and Zhengdong Wang and Zoubin Ghahramani},
      year={2025},
      eprint={2512.04797},
      archivePrefix={arXiv},
      primaryClass={cs.AI},
      url={https://arxiv.org/abs/2512.04797}, 
}

@article{10.1109/TPAMI.2024.3498346,
author = {Wang, Zhenyi and Yang, Enneng and Shen, Li and Huang, Heng},
title = {A Comprehensive Survey of Forgetting in Deep Learning Beyond Continual Learning},
year = {2025},
issue_date = {March 2025},
publisher = {IEEE Computer Society},
address = {USA},
volume = {47},
number = {3},
issn = {0162-8828},
url = {https://doi.org/10.1109/TPAMI.2024.3498346},
doi = {10.1109/TPAMI.2024.3498346},
journal = {IEEE Trans. Pattern Anal. Mach. Intell.},
month = mar,
pages = {1464–1483},
numpages = {20}
}

@INPROCEEDINGS {10655182,
author = { Zhu, Yichen and Ou, Zhicai and Mou, Xiaofeng and Tang, Jian },
booktitle = { 2024 IEEE/CVF Conference on Computer Vision and Pattern Recognition (CVPR) },
title = {{ Retrieval-Augmented Embodied Agents }},
year = {2024},
volume = {},
ISSN = {},
pages = {17985-17995},
doi = {10.1109/CVPR52733.2024.01703},
url = {https://doi.ieeecomputersociety.org/10.1109/CVPR52733.2024.01703},
publisher = {IEEE Computer Society},
address = {Los Alamitos, CA, USA},
month =Jun}

@article{raschka2020machine,
  title={Machine Learning in Python: Main developments and technology trends in data science, machine learning, and artificial intelligence},
  author={Raschka, Sebastian and Patterson, Joshua and Nolet, Corey},
  journal={arXiv preprint arXiv:2002.04803},
  year={2020}
}

@inproceedings{
lifshitz2023steve,
title={{STEVE}-1: A Generative Model for Text-to-Behavior in Minecraft},
author={Shalev Lifshitz and Keiran Paster and Harris Chan and Jimmy Ba and Sheila A. McIlraith},
booktitle={Thirty-seventh Conference on Neural Information Processing Systems},
year={2023},
url={https://openreview.net/forum?id=YkBDJWerKg}
}

@techreport{anthropic2026claudeopus5,
  title        = {Claude Opus 5 System Card},
  author       = {{Anthropic}},
  institution  = {Anthropic},
  year         = {2026},
  type         = {Technical Report},
  url          = {https://platform.claude.com/docs/en/resources/overview},
}
}

\newpage
\appendix
\crefalias{section}{appendix}
\crefalias{subsection}{appendix}

\section{\method{} Playthrough Examples}
\label{apx:playthrough}
Examples of how \method{} uses \textbf{dynamic bootstrapping} and \textbf{long-memory} to overcome roadblocks.

\begin{figure}[h]
  \centering
  \includegraphics[width=\linewidth]{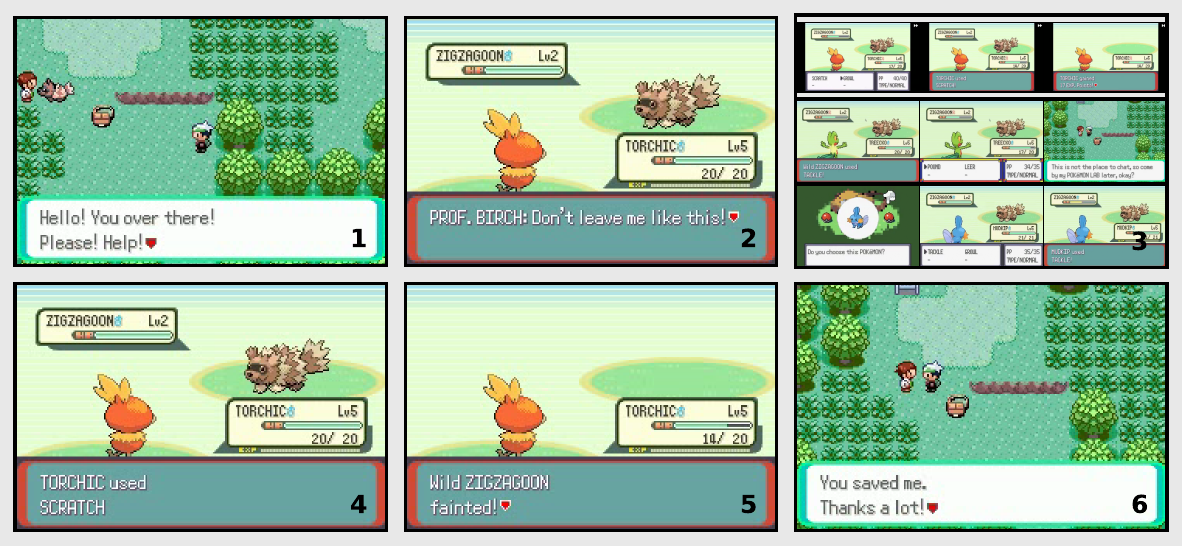}
  \vspace{-0.6cm}
  \caption{\textbf{Dynamic bootstrapping example.} To complete milestone 2, the player must rescue the Professor from a wild Zigzagoon \textbf{(Panel 1)}. To accomplish this, the player must use their starter \Pokemon{} to defeat the Zigzagoon in battle \textbf{(Panel 2)}. However, \method{}'s initial policy does not know how to use the battle interface to command their \Pokemon{}. After being stuck for $\Delta$ steps ($20$ minutes), \method{} \textbf{dynamically bootstraps}, and finds relevant examples of battling from \DI \textbf{(Panel 3)}. After updating its policy, \method{} is able to attack and defeat the Zigzagoon \textbf{(Panels 4 \& 5)}. After learning how to battle, \method{} is \textbf{unstuck} and continues on its journey \textbf{(Panel 6)}.}
\label{fig:bootstrap_example}
\vspace{-0.5cm}
\end{figure}

\begin{figure}[h]
  \centering
  \includegraphics[width=\linewidth]{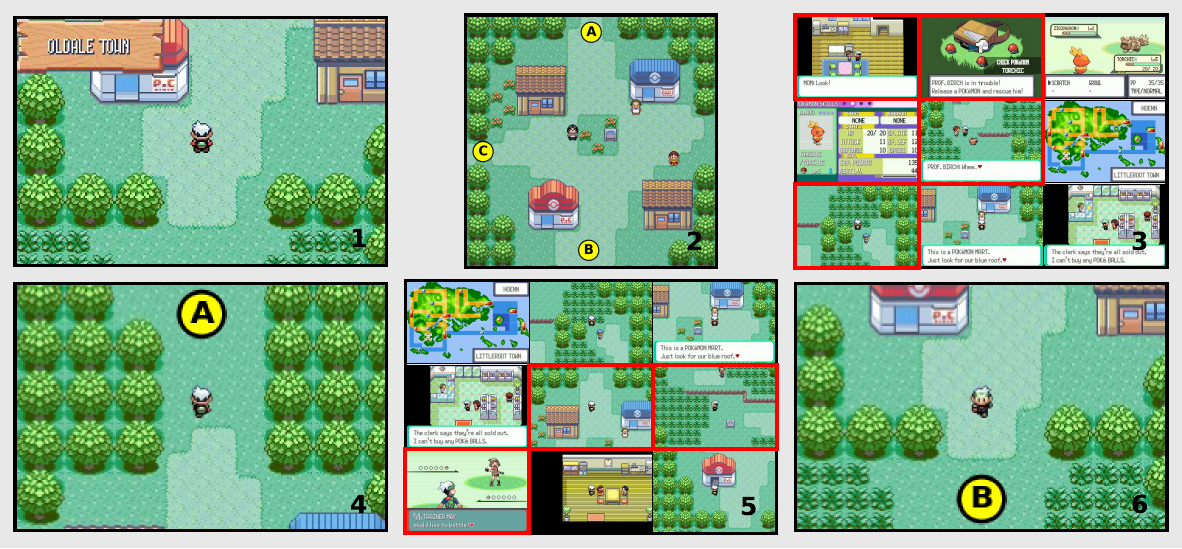}
  \vspace{-0.6cm}
  \caption{\textbf{Long-term memory example.} When the player arrives in Oldale town \textbf{(Panel 1)}, they are presented with 3 possible next paths. \textbf{Option \texttt{A}:} The player heads north to Route 103 to meet their rival, May. This is the correct choice if the player has just obtained their starter \Pokemon{} from the Professor and been tasked with bringing May back to the lab. \textbf{Option \texttt{B}:} The player has already met May, and should head back to the lab to meet the professor. Or \textbf{Option \texttt{C}:} The player has obtained the Pok\'edex at the lab from May and the Professor, and should proceed to Route 102. However, \textbf{short-term memory is insufficient for this decision}, as short-term observations are nearly identical. However, from inspecting the agent's \textbf{long-term memory (Panel 3)}, it is clear that the player has just obtained their starter \Pokemon{}, and therefore \method{} chooses \textbf{Option \texttt{A}} unambiguously \textbf{(Panel 4)}. After meeting May, the player's long-term memory has been updated with their encounter \textbf{(Panel 5)}, so it knows to choose \textbf{Option \texttt{B}} and head back to the lab \textbf{(Panel 6)}.}
\label{fig:mem_example}
\vspace{-0.3cm}
\end{figure}

\newpage

\section{ASH}
\label{apx:ash}
\input{algorithms/ash_v2}

\newpage
\section{Milestones}
\label{apx:milestones}

\begin{table}[!h]
\centering
\small
\begin{tabularx}{\linewidth}{|
  >{\centering\arraybackslash}m{0.18\linewidth}|
  >{\centering\arraybackslash}m{0.10\linewidth}|
  >{\raggedright\arraybackslash}X|
}
\hline
\textbf{Milestone} & \textbf{Icon} & \textbf{Description} \\
\hline
Home &
\includegraphics[width=1\linewidth]{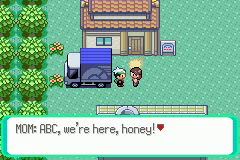} &
Before setting off on their journey, the player must talk to their mother and set the clock in their home.
\\
\hline
Save the professor &
\includegraphics[width=1\linewidth]{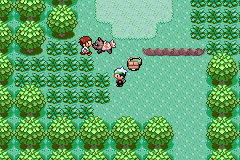} &
The player must obtain their starter \pokemon{} and save the professor from an encounter with a wild \pokemon{}.
\\
\hline
Oldale Town &
\includegraphics[width=1\linewidth]{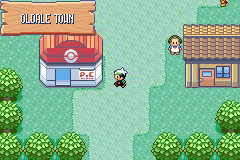} &
After progressing through Route 101, the player reaches the first town after their hometown. Oldale Town contains many important optional objectives such as healing your \pokemon{} at the \Pokemon{} Center and obtaining items at the Pok\'e Mart.
\\
\hline
Route 103 &
\includegraphics[width=1\linewidth]{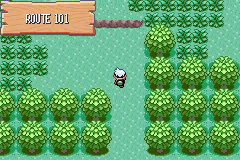} &
Route 103 is the area above Oldale Town. This milestone is achieved when the player first enters the route.
\\
\hline
Rival Battle &
\includegraphics[width=1\linewidth]{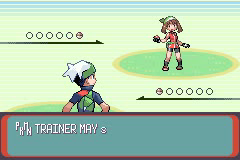} &
After meeting them on route 103, the player must defeat a trainer known as the "rival" in a \pokemon{} battle. This milestone is completed when the player successfully defeats their rival in a battle.
\\
\hline
Obtain Pok\'edex &
\includegraphics[width=1\linewidth]{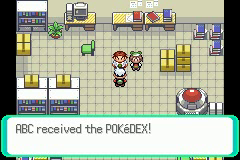} &
Upon defeating their rival, the player must backtrack to their hometown to obtain the Pok\'edex item from the professor. This milestone is completed when the player obtains the Pok\'edex.
\\
\hline
Route 102 &
\includegraphics[width=1\linewidth]{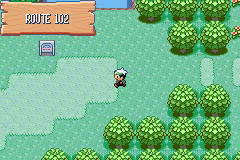} &
With the Pok\'edex in hand, the player can return to Oldale Town and continue on their journey through route 102. This milestone is achieved when the player first enters route 102.
\\
\hline
Arrive in Petalburg City &
\includegraphics[width=1\linewidth]{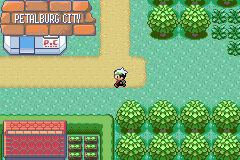} &
After traveling through Route 102, the player arrives in Petalburg City, where they meet key characters such as the gym leader (their father) and progress the main story.
\\
\hline
Defeat Team Aqua in Petalburg Woods &
\includegraphics[width=1\linewidth]{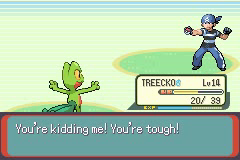} &
While passing through Petalburg Woods, the player encounters a Team Aqua member threatening a researcher. This milestone is completed when the player defeats the Team Aqua member in battle.
\\
\hline
Route 104 &
\includegraphics[width=1\linewidth]{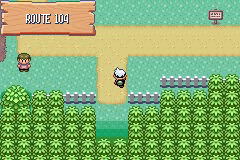} &
Route 104 connects Petalburg Woods to Rustboro City. This milestone is achieved when the player first enters Route 104 after exiting the forest.
\\
\hline
Rustboro City &
\includegraphics[width=1\linewidth]{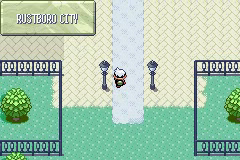} &
After traveling through Route 104, the player arrives in Rustboro City, home to the first Gym challenge and several important locations such as the Trainer's School and Devon Corporation.
\\
\hline
Defeat the Gym Leader &
\includegraphics[width=1\linewidth]{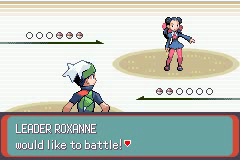} &
In Rustboro City, the player can battle the gym leader \emph{Roxanne}. Gym leaders are powerful trainers who act as bosses that the trainer must defeat to progress. The milestone is marked as completed \emph{after} the player wins the battle.
\\
\hline
\end{tabularx}
\caption{\label{tab:milestones}Milestones, icons and corresponding descriptions for the \Pokemon{} Emerald environment.}
\end{table}

\newpage

\begin{table}[!h]
\centering
\small
\begin{tabularx}{\linewidth}{|
  >{\centering\arraybackslash}m{0.18\linewidth}|
  >{\centering\arraybackslash}m{0.10\linewidth}|
  >{\raggedright\arraybackslash}X|
}
\hline
\textbf{Milestone} & \textbf{Icon} & \textbf{Description} \\
\hline
Obtain the sword from the smith &
\includegraphics[width=1\linewidth]{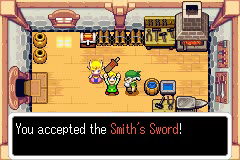} &
The player must speak to the smith to obtain a sword.
\\
\hline
Meet with Zelda &
\includegraphics[width=1\linewidth]{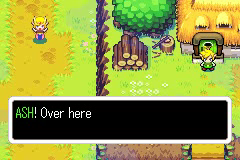} &
After obtaining the sword, the player must meet up with Zelda to head to the festival. This milestone is marked as completed when the player `groups up' with Zelda and she follows the player to the festival.
\\
\hline
Obtain the small shield &
\includegraphics[width=1\linewidth]{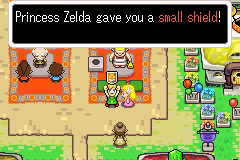} &
At the festival, the player must win the small shield item as a prize in a contest. The milestone is completed when the player obtains the shield.
\\
\hline
Reflect the seed projectile with the shield &
\includegraphics[width=1\linewidth]{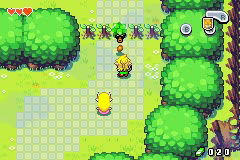} &
After leaving the festival, the player encounters a \emph{Deku Scrub} who is firing seeds and blocking the path. To progress, the player must use their shield to deflect a seed back at the Deku Scrub to stun it. 
\\
\hline
Enter Hyrule Castle &
\includegraphics[width=1\linewidth]{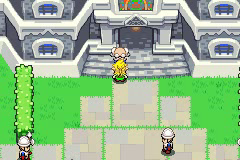} &
The player arrives at the entrance to Hyrule Castle. This milestone is achieved when the player reaches the castle exterior.
\\
\hline
Talk with the king &
\includegraphics[width=1\linewidth]{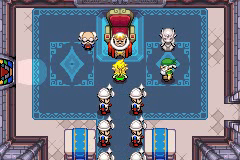} &
The player enters the throne room and has an audience with the King of Hyrule.
\\
\hline
Leave the castle &
\includegraphics[width=1\linewidth]{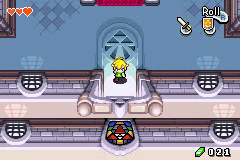} &
After speaking with the king, the player must leave the castle. This milestone is marked as completed once the player exits the castle. We allow the player to use any of the castle exits for this milestone.
\\
\hline
Leave the palace grounds &
\includegraphics[width=1\linewidth]{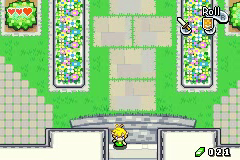} &
This milestone is completed when the player leaves the palace grounds area.
\\
\hline
Enter Lon Lon Ranch &
\includegraphics[width=1\linewidth]{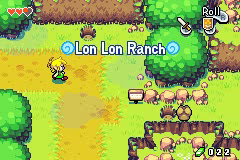} &
To get to the Minish Forest, the player must go through the \emph{Lon Lon Ranch} zone, this milestone is completed when the player first enters this new area.
\\
\hline
Enter the Minish Woods &
\includegraphics[width=1\linewidth]{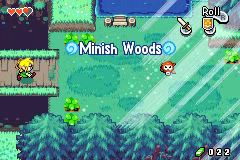} &
After navigating through Lon Lon Ranch, the player can enter the Minish Forest, an important location for the player's quest. This milestone is completed when the player enters the forest.
\\
\hline
Obtain the Minish Cap &
\includegraphics[width=1\linewidth]{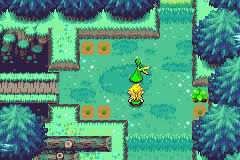} &
In the Minish Forest, the player obtains the \emph{Minish Cap}. A new item that allows the player to shrink in size.
When the player places the Minish Cap on their head, this milestone is completed.
\\
\hline
Solve the Minish Cap Puzzle &
\includegraphics[width=1\linewidth]{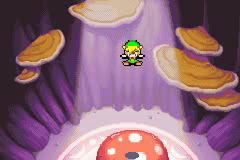} &
Using the new shrinking game mechanic (from the Minish Cap), the player must solve a puzzle that requires using the Minish Cap to fit through narrow spaces
(such as the interior of a log). This milestone is completed when the player finishes the puzzle.
\\
\hline
\end{tabularx}
\caption{\label{tab:lz_milestones}Milestones, icons and corresponding descriptions for the Legend of Zelda environment.}
\end{table}
\newpage

\textbf{Milestone distribution in human walkthroughs}.
Our evaluations are sensitive to milestone choice. If our method made marginally more progress than the next-best baseline, and there were a disproportionate number of milestones in the slice of gameplay separating our method from the next-best, this would make our method look substantially better than it actually was. To evaluate this property of our milestones, we take 5 online walkthroughs of Pokemon Emerald and report the average time at which the players completed each milestone. We report results in \Cref{tab:milestone_timing}.

\begin{table}[h]
\centering
\small
\begin{tabular}{clr}
\toprule
\# & Milestone & Avg.\ completion time (min) \\
\midrule
1  & Home                                 & $7.91$ \\
2  & Save the professor                   & $15.27$ \\
3  & Oldale Town                          & $22.28$ \\
4  & Route 103                            & $24.09$ \\
5  & Rival Battle                         & $26.41$ \\
6  & Obtain Pok\'edex                     & $28.72$ \\
7  & Route 102                            & $36.68$ \\
8  & Arrive in Petalburg City             & $49.43$ \\
9  & Defeat Team Aqua in Petalburg Woods  & $85.47$ \\
10 & Route 104                            & $89.74$ \\
11 & Rustboro City                        & $101.04$ \\
12 & Defeat the Gym Leader                & $149.51$ \\
\bottomrule
\end{tabular}
\caption{\label{tab:milestone_timing}Average completion time per milestone across 5 human walkthroughs of \Pokemon{} Emerald.}
\end{table}

We see that the milestones are more concentrated towards the beginning of the game, suggesting that differences between average milestones completed are more significant when the absolute number is higher. \method{} completes an average of 11.2 \Pokemon{} milestones compared to the next best baseline which completes 9. This is equivalent to $\sim 15$ minutes of human play.

\section{Background on Environments}
\label{apx:pokemon}

\textbf{\Pokemon{} Emerald} is a third-generation \emph{\Pokemon{}} role-playing game for the Game Boy Advance (GBA). The game follows a largely linear-but-explorable progression: the player travels between towns and routes, completes story events, captures and trains creatures ("\Pokemon{}"), and defeats a sequence of gym leaders to earn badges, culminating in the Elite Four and Champion battles.

Gameplay consists of real-time overworld navigation (movement, interactions with NPCs, and menu-driven inventory/team management) interleaved with turn-based battles. Battles require selecting discrete actions (moves, items, switching Pok\'emon, or attempting capture) under uncertainty due to hidden information (e.g., opponent move choice, variable damage rolls, status effects). Progress is gated by narrative triggers and abilities acquired over time (e.g., key items and field moves), producing long-horizon dependencies, sparse success signals, and frequent context switches between exploration, dialogue, menus, and combat. \citet{Bulbapedia_2025} provides a comprehensive overview of progression, game mechanics and storyline.

\Pokemon{} Emerald is a demanding test of long-horizon visuomotor intelligence: it must map high-dimensional visual observations to low-level button sequences while coping with partial observability, delayed consequences, and sparse success signals.
The outcomes of many actions are unclear over even long time horizons. For example, your starter \pokemon{} choice may be a crucial factor in winning a battle dozens of hours into a playthrough.

\textbf{Legend of Zelda: The Minish Cap} is a Game Boy Advance game that combines real-time combat, puzzle solving, and RPG mechanics.
The player controls Link, a young hero tasked with rescuing Princess Zelda after she is turned to stone by the evil sorcerer Vaati during the annual Picori Festival.
To restore Zelda and prevent Vaati from unleashing further chaos upon the kingdom of Hyrule, Link embarks on a journey to reforge the sacred Picori Blade.

The core gameplay loop consists of exploring a diverse set of zones collecting items to save the princess, defeating enemies with the player's sword and shield, and solving increasingly difficult puzzles using the game's RPG mechanics.
The game's central mechanic is the ability to use the Minish Cap to shrink Link.
This mechanic (in conjunction with other game mechanics) is combined with environmental puzzles to create settings where the player must reason how to use Link's ability to grow and shrink to progress through novel areas and terrains.
This mechanic is also central to the combat in the game: when Link shrinks enemies that were once trivial become large and imposing.
\citet{minishcap} provides a detailed walkthrough of the game, including areas, game mechanics, and the storyline of Link's quest.

Legend of Zelda: The Minish Cap is a fascinating environment for several reasons: (1) As with our previous environment, it is long-horizon, with progression spanning many hours and distinct areas. (2) \emph{Progressing requires that the player demonstrates a variety of skills.} The Legend of Zelda is not a game where the player can simply ``walk to the finish line,'' progressing through each zone requires demonstrating \emph{escalating levels of mastery} over the game's core combat mechanics (centered around a sword and shield), the ability to traverse movement challenges, and utilize newly-introduced game mechanics to solve puzzles. (3) Lastly, \emph{the environment rewards open-ended exploration}. As the player is tasked with a series of progressively harder challenges, they can explore the environment to find power-ups that make these challenges easier. For example, the player can obtain \emph{heart containers} which increase the amount of damage the player can take without dying in combat.

\section{Extended Experiments}

\subsection{Average Milestone Results}
\label{apx:final_milestones}

In \Cref{tab:final_milestones}, we report the average number of milestones reached (out of 12) in both environments at the end of the 8-hour evaluation.

\begin{table}[h]
\centering
\small
\begin{tabular}{lcc}
\toprule
Method & \Pokemon{} Emerald & Legend of Zelda \\
\midrule
\method{} & $\mathbf{11.17 \pm 0.83}$ & $\mathbf{9.92 \pm 1.51}$ \\
Voyager+Opus 5   & $9.25 \pm 1.71$           & $7.83 \pm 2.04$ \\
BC        & $6.50 \pm 1.31$           & $6.00 \pm 0.00$ \\
VPT       & $5.50 \pm 1.62$           & $5.92 \pm 0.29$ \\
RAEA      & $2.58 \pm 0.51$           & $2.00 \pm 0.00$ \\
\bottomrule
\end{tabular}
\caption{\label{tab:final_milestones}Average milestones reached (out of 12) at the end of the 8-hour evaluation, reported as mean $\pm$ one standard deviation.}
\end{table}

\subsection{Sensitivity to the Scale and Quality of Internet Video}
\label{apx:corpus}

\textbf{Corpus size.} To measure how sensitive \method{} is to the scale of our dataset, we ablate using 10\%, 50\% and 100\% of our corpora in our Pokemon Emerald environment. For each dataset size, we average over 4 runs, each 4 hours in length.

\begin{table}[h]
\centering
\small
\begin{tabular}{lc}
\toprule
Corpus size & Avg.\ milestones \\
\midrule
$10\%$  & $6.75$ \\
$50\%$  & $7.75$ \\
$100\%$ & $8.25$ \\
\bottomrule
\end{tabular}
\caption{\label{tab:corpus_size}Effect of internet corpus size on \method{} in \Pokemon{} Emerald. Each value is an average over 4 runs of 4 hours.}
\end{table}

We report average milestones completed for each dataset size in \Cref{tab:corpus_size}. The average milestones completed drops by ~1.5 with a 90\% reduction in dataset size. Therefore, while \method{} does benefit from the availability of data, it can still function in settings with substantially less data.

\textbf{Corpus quality.} To evaluate the quality of our data, we sample 200 videos at random from our internet corpus. For each video, we record if it (1) is from a wrong or modded version of the game, (2) has major editing artifacts (custom overlays, facecams, etc.), (3) has poor video quality (blurry due to low resolution or choppy due to low framerate). If any of these are true, we consider the video to be “dirty”. We report the results of this analysis in \Cref{tab:corpus_quality}.

\begin{table}[h]
\centering
\small
\begin{tabular}{lc}
\toprule
Issue & \% of videos \\
\midrule
Wrong or modded game version & $16.5$ \\
Major editing artifacts       & $35.5$ \\
Poor video quality            & $19.0$ \\
\midrule
Dirty (any of the above)      & $52.5$ \\
\bottomrule
\end{tabular}
\caption{\label{tab:corpus_quality}Manual quality audit of 200 videos sampled uniformly at random from \DI{} for \Pokemon{} Emerald. The three categories are not mutually exclusive.}
\end{table}

We see that more than half (52.5\%) of the corpus is dirty. ASH’s ability to extract high quality supervision from this dirty corpus is one of its unique advantages. When VPT and BC baselines were trained directly on this data, they achieved 0 milestones. This indicates that the ASH algorithm as a whole is more robust to noisy data than other methods.

\subsection{Catastrophic Forgetting in the IDM}
\label{apx:idm_forgetting}

\Cref{exp:idm} shows that IDM accuracy rises from $60\%$ to $81\%$ over 8 bootstraps. Each bootstrap trains the IDM on data that \method{} retrieves for the agent's \emph{current} roadblock. The IDM could therefore learn the dynamics of the later game and forget the dynamics of the earlier game. To evaluate if the IDM module forgets previous levels, we evaluate \Pokemon{} Emerald IDM checkpoints from successive bootstraps on data collected during the first bootstrap, and report movement class prediction accuracy in \Cref{tab:idm_forgetting}.

\begin{table}[h]
\centering
\small
\begin{tabular}{lcccccccc}
\toprule
IDM checkpoint (bootstrap \#) & 1 & 2 & 3 & 4 & 5 & 6 & 7 & 8 \\
\midrule
Accuracy on bootstrap-1 data (\%) & $72.5$ & $78.6$ & $74.0$ & $76.5$ & $78.3$ & $68.9$ & $77.1$ & $74.5$ \\
\bottomrule
\end{tabular}
\caption{\label{tab:idm_forgetting}Movement class accuracy of successive IDM checkpoints. We evaluate every checkpoint on held-out data from the first bootstrap.}
\end{table}

Accuracy on early-game data stays largely flat across the 8 bootstraps, ranging from $68.9\%$ to $78.6\%$ with no downwards trend. The IDM therefore does not forget the earlier parts of the game while it improves.

\subsection{Sensitivity to the Stuck-Timer Threshold $\Delta$}
\label{apx:delta}

Dynamic bootstrapping has one hyperparameter: the stuck-timer threshold $\Delta$, the number of steps that an agent may run without a new key moment before \method{} bootstraps. To evaluate \method{}'s sensitivity to the stuck threshold $\Delta$, we perform an ablation. For each value of $\Delta$, we report average milestones completed for 4 runs, each 96 GPU hours in length (\Cref{tab:delta}).

\begin{table}[h]
\centering
\small
\begin{tabular}{lcccc}
\toprule
$\Delta$ (minutes) & 5 & 10 & 20 & 40 \\
\midrule
Avg.\ milestones & $4.00$ & $6.50$ & $6.25$ & $7.00$ \\
\bottomrule
\end{tabular}
\caption{\label{tab:delta}Sensitivity to the stuck-timer threshold $\Delta$ in \Pokemon{} Emerald. Each value is an average over 4 runs of 96 GPU hours of online training. We use $\Delta = 20$ minutes throughout the main paper.}
\end{table}

We see that at very low $\Delta$ value (5 minutes), the agent does not spend enough time exploring, and bootstraps too frequently. However, above this value, ASH’s performance does not seem very sensitive to $\Delta$.

\subsection{General Skill Acquisition}
\label{apx:battle}

A method that learns from walkthrough video could simply repeat actions from the trajectory it saw, rather than learning general skills. In \Pokemon{} Emerald, the core skill is \Pokemon{} battling: when the player encounters enemies they must use their team of \Pokemon{} to defeat them. To measure if the agent is learning this skill, we report the average rate of successful battles in successive bootstraps \Cref{tab:battle}.

\begin{table}[h]
\centering
\small
\begin{tabular}{lcccccc}
\toprule
Bootstrap & 0 & 1 & 2 & 3 & 4 & 5 \\
\midrule
Battles won (\%) & $0.0$ & $40.0$ & $62.5$ & $80.7$ & $92.1$ & $95.4$ \\
\bottomrule
\end{tabular}
\caption{\label{tab:battle}Success rate in randomized battles across successive bootstraps in \Pokemon{} Emerald.}
\end{table}

We observe that the rate of success in these random encounters increases over bootstraps. This suggests the agent is learning to battle - which is a general skill, and not simply to copy expert demonstrations.

\section{Hyperparameters and Training Settings}
\label{apx:settings}
\begin{table}[!h]
\centering
\small

\renewcommand{\arraystretch}{1.25} 
\setlength{\extrarowheight}{1.5pt} 

\begin{tabular}{|
  >{\centering\arraybackslash}m{0.50\linewidth}|
  >{\centering\arraybackslash}p{0.50\linewidth}|
}
\hline
\textbf{Hyperparameter} & \textbf{Setting} \\
\hline
Optimizer & AdamW~\citep{loshchilov2019decoupledweightdecayregularization} \\
\hline
Learning Rate & $5 \times 10^{-5}$ \\
\hline
LR Scheduling & Linear \\
\hline
Warm-up Ratio & $5\%$ \\
\hline
Weight Decay & $1 \times 10^{-5}$ \\
\hline
Max Epochs & $3$ \\
\hline
Batch Size & $4$ \\
\hline
Trajectories ($N$) & $4$ \\
\hline
Retrieved Videos per Agent $k$ (for training $\pi$) & $100$ \\
\hline
Sampling Temperature & $1$ \\
\hline
Seconds between actions & $0.25$ \\
\hline
Observations in Short-term Memory ($\shortlen$) & $64$ \\
\hline
Seconds between observations & $0.5$ \\
\hline
$\mathcal{T}$ hidden dimension & $1024$ \\
\hline
$\phi$ hidden dimension & $768$ \\
\hline
HDBSCAN minimum samples & $30$ \\
\hline
\end{tabular}

\caption{\label{tab:hp}Hyperparameters and Settings.}
\end{table}

\newpage
\section{Notation Glossary}
\label{apx:notation}

\begin{table}[h]
\centering
\small
\begin{tabular}{cl}
\toprule
\textbf{Symbol} & \textbf{Description} \\
\midrule
\multicolumn{2}{l}{\textit{Spaces and environment}} \\
$\mathcal{E}$ & Environment \\
$\mathcal{A}$ & Action space \\
$\mathcal{O}$ & Observation space \\
$a \in \mathcal{A}$ & Action \\
$o \in \mathcal{O}$ & Observation \\
\midrule
\multicolumn{2}{l}{\textit{Trajectories}} \\
$\tau$ & Observation-action trajectory $(o_1, a_1, \dots, o_T, a_T)$ \\
$\obs$ & Observation-only trajectory $(o_1, \dots, o_T)$ \\
$\tau_n$ & Trajectory of agent $n$ \\
$\tau_{\oneToN}$ & Trajectories all $N$ agents: $(\tau_1,\dots,\tau_N)$ \\
$\sample$ & Short-term sample $o_{t:t+\shortlen}$ \\
\midrule
\multicolumn{2}{l}{\textit{Policy and models}} \\
$\pi \sim p_\theta(a_{t+1} \mid \cdot)$ & Policy \\
$\phi$ & Image tokenizer (SigLIP) \\
$\mathcal{T}$ & Causal transformer backbone \\
$\text{IDM}$ & Inverse dynamics model \\
$\tilde{\mA}$ & Pseudo-action labels from IDM \\
$\mW_o$ & Output projection matrix \\
$\mH$ & Hidden states of $\mathcal{T}$ \\
$\mathcal{L}$ & Training loss \\
\midrule
\multicolumn{2}{l}{\textit{Memory}} \\
$\shortlen$ & Short-term memory window size (number of observations) \\
$\longlen$ & Long-term memory window size (number of key moments) \\
$\lmem$ & Long-term memory (most recent $\longlen$ key moments) \\
$\shorttok$ & Short-term memory tokens from $\phi$ \\
$\longtok$ & Long-term memory tokens from $\phi$ \\
$\mathcal{K}(\obs, t)$ & Key-moment classifier \\
$\mathcal{M}_n$ & Memory bank of agent $n$ \\
$\Delta$ & Stuck-timer threshold (steps before bootstrap) \\
\midrule
\multicolumn{2}{l}{\textit{Data and retrieval}} \\
$\DI$ & Internet video dataset \\
$\DR$ & Retrieved video dataset \\
$\Dpi$ & Policy update dataset \\
$\mE \in \mathbb{R}^{e \times d}$ & Per-video DINOv2 embedding matrix \\
$\mQ$ & Per-trajectory query embedding matrix \\
$\mS = \mQ\mE^\top$ & Similarity matrix \\
$\retlen$ & Retrieval window size \\
$k$ & Number of retrieved videos per agent \\
$c_{\min}$ & Minimum unique trajectories in a cluster \\
\midrule
\multicolumn{2}{l}{\textit{\method{}}} \\
$N$ & Number of parallel agents \\
\bottomrule
\end{tabular}

\caption{Notation glossary.}
\label{tab:notation}
\end{table}

\newpage

\section{Search Keywords}
\label{apx:keyword}

\Cref{tab:search_strings} lists the search terms used to create our internet dataset.
To ensure that we index multi-part walkthroughs, we replace \{n\} with the values 1-80.

\begin{table}[h]
\centering
\small
\begin{tabularx}{\linewidth}{|>{\raggedright\arraybackslash}X|>{\raggedright\arraybackslash}X|}
\hline
\textbf{Pok\'emon Emerald} & \textbf{Legend of Zelda: The Minish Cap} \\
\hline
pokemon emerald full walkthrough & the minish cap full walkthrough \\
pokemon emerald longplay & legend of zelda the minish cap full walkthrough \\
pokemon emerald walkthrough & the legend of zelda the minish cap full walkthrough \\
pokemon emerald gameplay & the minish cap longplay \\
pokemon emerald playthrough & legend of zelda the minish cap longplay \\
pokemon emerald full game & the minish cap walkthrough \\
pokemon emerald full gameplay & legend of zelda the minish cap walkthrough \\
pokemon emerald all parts & the minish cap gameplay \\
pokemon emerald complete & legend of zelda the minish cap gameplay \\
pokemon emerald full version & the minish cap playthrough \\
pokemon emerald no commentary & legend of zelda the minish cap playthrough \\
pokemon emerald 100 percent & the minish cap full game \\
pokemon emerald part \{n\} & legend of zelda the minish cap full game \\
pokemon emerald longplay part \{n\} & the minish cap full gameplay \\
pokemon emerald let's play part \{n\} & legend of zelda the minish cap full gameplay \\
pokemon emerald episode \{n\} & the minish cap all parts \\
pokemon esmeralda gu\'ia completa & legend of zelda the minish cap all parts \\
pokemon esmeralda longplay & the minish cap complete \\
pokemon esmeralda gu\'ia & legend of zelda the minish cap complete \\
pokemon esmeralda gameplay & the minish cap full version \\
pokemon esmeralda recorrido & legend of zelda the minish cap full version \\
pokemon esmeralda juego completo & the minish cap no commentary \\
pokemon esmeralda gameplay completo & legend of zelda the minish cap no commentary \\
pokemon esmeralda todas las partes & the minish cap 100 percent \\
pokemon esmeralda completo & legend of zelda the minish cap 100 percent \\
pokemon esmeralda versi\'on completa & the minish cap 100\% \\
pokemon esmeralda sin comentarios & legend of zelda the minish cap 100\% \\
pokemon esmeralda 100\% & the minish cap part \{n\} \\
pokemon esmeralda parte \{n\} & the minish cap longplay part \{n\} \\
\hline
\end{tabularx}
\caption{\label{tab:search_strings}Search terms used for creating internet dataset.}
\end{table}

\section{Broader Impact}
\label{impacts}
\textbf{Positive.} By enabling agents to acquire new skills from widely available observational data, methods such as ASH could reduce the cost of developing systems for domains including robotics, assistive technologies, simulation-based training, and autonomous exploration, where hand-engineered rewards or large labeled datasets are often impractical. More broadly, scalable long-horizon learning may accelerate the development of adaptive systems that can operate in complex real-world environments.

\textbf{Negative.} Systems capable of autonomously acquiring behaviors from internet-scale demonstrations may inherit biases, unsafe behaviors, or undesirable strategies present in online data. In addition, increasingly capable agents could be applied in ways that reduce human oversight in safety-critical settings or be adopted for harmful applications. Our experiments are limited to video game environments, which provide a safe and controlled testbed, but future deployment in physical domains will require robust safeguards.

\end{document}